\documentclass[10pt,journal,cspaper,compsoc]{IEEEtran}

\ifCLASSOPTIONcompsoc
\else
\fi

\ifCLASSINFOpdf
\else
\fi

\usepackage{amssymb}
\usepackage{amsmath}
\usepackage{algorithm}
\usepackage{algpseudocode}
\usepackage{threeparttable}

\newcommand{\xv}{\mathbf{x}}

\newcommand{\yv}{\mathbf{y}}

\newcommand{\zv}{\mathbf{z}}
\newcommand{\Zv}{\mathbf{Z}}

\newcommand{\hv}{\mathbf{h}}

\newcommand{\Cv}{\mathbf{C}}

\newcommand{\pv}{\mathbf{p}}

\newcommand{\wv}{\mathbf{w}}

\newcommand{\argmax}{\operatornamewithlimits{argmax}}

\newcommand{\ud}{\mathrm{d}}

\newcommand{\etav}{\boldsymbol \eta}
\newcommand{\zetav}{\boldsymbol \zeta}
\newcommand{\epsilonv}{\boldsymbol \epsilon}
\newcommand{\muv}{\boldsymbol \mu}

\newcommand{\psiv}{\boldsymbol \psi}
\newcommand{\phiv}{\boldsymbol \phi}

\newcommand{\thetav}{\boldsymbol \theta}
\newcommand{\Thetav}{\boldsymbol \Theta}
\newcommand{\alphav}{\boldsymbol \alpha}
\newcommand{\betav}{\boldsymbol \beta}

\newcommand{\ep}{\mathbb{E}}
\newcommand{\prob}{\mathcal{P}}

\newcommand{\KL}{\mathrm{KL}}

\newcommand{\M}{\mathrm{Mult}}

\newcommand{\model}{\mathbf{M}}

\newcommand{\data}{\mathcal{D}}
\def\indicator{{\mathbb I}}

\usepackage{enumerate}
\usepackage{epsf}
\usepackage{multirow}
\usepackage{wrapfig}
\usepackage{graphicx}
\usepackage{subfigure} 
\usepackage[english]{babel}

\usepackage{color}
\newcommand{\junz}[1]{{\color{blue}{\bf\sf [Jun: #1]}}}
\newcommand{\jun}[1]{{\color{blue}{\bf\sf #1}}}

\begin{document}
%
\title{Big Learning with Bayesian Methods}

\author{Jun Zhu, ~Jianfei Chen,~ Wenbo Hu,~Bo Zhang
\IEEEcompsocitemizethanks{\IEEEcompsocthanksitem J. Zhu, J. Chen, W. Hu, and B. Zhang are with TNList Lab; State Key Lab for Intelligent Technology and Systems; Department of 
Computer Science and Technology, Tsinghua University, Beijing, 100084 China. 
Email: \{dcszj,~dcszb\}@tsinghua.edu.cn; \{chenjian14,~hwb13\}@mails.tsinghua.edu.cn}
\thanks{}}

%

\IEEEcompsoctitleabstractindextext{%
\begin{abstract}
The explosive growth in data volume and the availability of cheap computing resources have
sparked increasing interest in Big learning, an emerging subfield that studies
scalable machine learning algorithms, systems, and applications with Big Data.
Bayesian methods represent one important class of statistical methods for machine
learning, with substantial recent developments on adaptive, flexible and
scalable Bayesian learning.  This article provides a survey of the recent
advances in Big learning with Bayesian methods, termed Big Bayesian Learning,
including nonparametric Bayesian methods for adaptively inferring model
complexity, regularized Bayesian inference for improving the flexibility via
posterior regularization, and scalable algorithms and systems based on
stochastic subsampling and distributed computing for dealing with large-scale
applications. We also provide various new perspectives on the large-scale Bayesian
modeling and inference.

\end{abstract}

}

\maketitle

\IEEEdisplaynotcompsoctitleabstractindextext

%
\IEEEpeerreviewmaketitle

\section{Introduction}

%
%
%
\IEEEPARstart{W}{e} live in an era of Big Data, where science, engineering and
technology are producing massive data streams, with petabyte and exabyte scales
becoming increasingly common~\cite{Brumfiel:2011,Doctorow:2008,Reichman:2011}.
Besides the explosive growth in volume, Big Data also has high velocity,
high variety, and high uncertainty.
These complex data streams require ever-increasing processing speeds, economical
storage, and timely response for decision making in highly uncertain
environments, and have raised various challenges to conventional data analysis~\cite{Fan:NSR13}.

With the primary goal of building intelligent systems that automatically improve
from experiences, machine learning (ML) is becoming an increasingly important
field to tackle the big data challenges~\cite{Mitchell:1997}, with an emerging
field of {\it Big Learning}, which covers theories, algorithms and systems on
addressing big data problems.




\subsection{Big Learning Challenges}

In big data era, machine learning needs to deal with the challenges of learning
from complex situations with {\it large $N$, large $P$, large $L$}, and {\it
large $M$}, where $N$ is the data size, $P$ is the feature dimension, $L$ is the
number of tasks, and $M$ is the model size. Given that $N$ is obvious, we
explain the other factors below.


{\bf Large $P$}: with the development of Internet, data sets with ultrahigh
dimensionality have emerged, such as the spam filtering data with trillion
features~\cite{Weinberger:2009} and the even higher-dimensional feature space
via explicit kernel mapping~\cite{Tan:jmlr14}. Note that whether a learning
problem is high-dimensional depends on the ratio between $P$ and $N$. Many
scientific problems with $P \gg N$ impose great challenges on learning, calling
for effective regularization techniques to avoid overfitting and select salient
features~\cite{Fan:NSR13}.

{\bf Large $L$}: many tasks involve classifying text or images into tens of
thousands or millions of categories. For example, the ImageNet~\cite{ImageNet}
database consists of more than 14 millions of web images from 21 thousands of
concepts, while with the goal of providing on average 1,000 images for each of
100+ thousands of concepts (or synsets) in WordNet; and the LSHTC text
classification challenge 2014 aims to classify Wikipedia documents into one of
325,056 categories~\cite{LSHTC}. Often, these categories are organized in a
graph, e.g., the tree structure in ImageNet and the DAG (directed acyclic graph)
structure in LSHTC, which can be explored for better
learning~\cite{Bengio:nips10,Deng:nips11}.

{\bf Large $M$}: with the availability of massive data, models with millions or
billions of parameters are becoming common. Significant progress has been made
on learning deep models, which have multiple layers of non-linearities allowing
them to extract multi-grained representations of data, with successful
applications in computer vision, speech recognition, and natural language
processing. Such models include neural networks~\cite{Hinton:2012},
auto-encoders~\cite{Vincent:2008,Quoc:2012}, and probabilistic generative
models~\cite{Salakhutdinov:2009,Rezende:2014}.


\subsection{Big Bayesian Learning}

Though Bayesian methods have been widely used in machine learning and many other areas, skepticism often arises
when we talking about Bayesian methods for big data~\cite{Jordan:2011}.
Practitioners also criticize that Bayesian methods are often too slow for even
small-scaled problems, owning to many factors such as the non-conjugacy models
with intractable integrals. Nevertheless,
Bayesian methods have several advantages on dealing with:
\begin{enumerate}
\item {\bf Uncertainty}: our world is an uncertain place because of physical
randomness, incomplete knowledge, ambiguities and contradictions. Bayesian
methods provide a principled theory for combining prior knowledge and uncertain
evidence to make sophisticated inference of hidden factors and predictions.
\item {\bf Flexibility}: Bayesian methods are conceptually simple and flexible.
Hierarchical Bayesian modeling offers a flexible tool for characterizing
uncertainty, missing values, latent structures, and more. Moreover, regularized
Bayesian inference (RegBayes)~\cite{Zhu:regBayes14} further augments the flexibility
by introducing an extra dimension (i.e., a posterior regularization term)
to incorporate domain knowledge or to optimize a learning objective. Finally, 
there exist very flexible algorithms (e.g., Markov Chain Monte Carlo) to perform posterior inference.
\item {\bf Adaptivity}: The dynamics and uncertainty of Big Data require that
our models should be adaptive when the learning scenarios change. Nonparametric
Bayesian methods provide elegant tools to deal with situations in which
phenomena continue to emerge as data are collected~\cite{Hjort:2010}. Moreover,
the Bayesian updating rule and its variants are sequential in nature and
suitable for dealing with big data streams.
\item {\bf Overfitting}: Although the data volume grows exponentially, the
predictive information grows slower than the amount of Shannon
information~\cite{Bialek:2001}, while our models are becoming increasingly large
by leveraging powerful computers, such as the deep networks with billions of
parameters. It implies that our models are increasing their capacity faster than
the amount of information that we need to fill them with, therefore causing
serious overfitting problems that call for effective
regularization~\cite{Srivastava:2014}.
\end{enumerate}

Therefore, Bayesian methods are becoming increasingly relevant in the big data
era~\cite{Welling:2014} to protect high capacity models against overfitting, and
to allow models adaptively updating their capacity. However, the application of
Bayesian methods to big data problems runs into a computational bottleneck that
needs to be addressed with new (approximate) inference methods.
This article aims to provide a literature survey of the recent advances in big
learning with Bayesian methods, including the basic concepts of Bayesian inference, nonparametric Bayesian methods,
regularized Bayesian inference, scalable inference algorithms and systems based
on stochastic subsampling
and distributed computing.  

It is useful to note that our review is no way exhaustive. We select the materials to 
make it self-contained and technically rigorous. As data analysis is becoming an essential function 
in many scientific and engineering areas, this article should be of broad interest to the audiences 
who are dealing with data, especially those who are using statistical tools. 

%
%

\section{Basics of Bayesian Methods}


The general blueprint of Bayesian data analysis~\cite{Gelman:2013} is that a
Bayesian model expresses a generative process of the data that includes hidden
variables, under some statistical assumptions. The process specifies a joint
probability distribution of the hidden and observed random variables. Given a
set of observed data, data analysis is performed by {\it posterior inference},
which computes the conditional distribution of the hidden variables given the
observed data. This section reviews the basic concepts and algorithms of Bayesian inference.

\subsection{Bayes' Theorem}

At the core of Bayesian methods is Bayes' theorem (a.k.a Bayes' rule). Let
$\Thetav$ be the model parameters and $\data$ be the given data set. The
Bayesian posterior distribution is
\begin{eqnarray}
p(\Thetav | \data) = \frac{p_0(\Thetav) p(\data | \Thetav)}{ p(\data)},
\end{eqnarray}
where $p_0(\cdot)$ is a prior distribution, chosen before seeing any data;
$p(\data | \Thetav)$ is the assumed likelihood model; and $p(\data) = \int
p_0(\Thetav) p(\data | \Thetav) \ud \Thetav$ is the marginal likelihood (or
evidence), often involving an intractable integration problem that requires
approximate inference as detailed below. The year 2013 marks the 250th
anniversary of Thomas Bayes' essay on how humans can sequentially learn from
experience, steadily updating their beliefs as more data become
available~\cite{Efron:science13}.

A useful variational formulation of Bayes' rule is
\begin{eqnarray}\label{eq:varBayes}
\min_{q(\Thetav) \in \prob} \KL( q(\Thetav) \Vert p_0(\Thetav) ) - \ep_q[ \log p(\data | \Thetav) ],
\end{eqnarray}
where $\prob$ is the space of all distributions that make the objective well-defined.
It can be shown that the optimum solution to (\ref{eq:varBayes}) is identical to
the Bayesian posterior. In fact, if we add the constant term $\log p(\data)$,
the problem is equivalent to minimizing the KL-divergence between $q(\Thetav)$
and the Bayesian posterior $p(\Thetav | \data)$, which is non-negative and takes
$0$ if and only if $q$ equals to $p(\Thetav | \data)$. The variational
interpretation is significant in two aspects: (1) it provides a basis for
variational Bayes methods; and (2) it provides a starting point to make Bayesian
methods more flexible by incorporating a rich set of posterior constraints. We
will make these clear soon later.

It is noteworthy that $q(\Thetav)$ represents the density of a general post-data
posterior in the sense of \cite[pp.15]{Ghosh:book2003}, not necessarily
corresponding to a Bayesian posterior induced by Bayes' rule. {As we shall see
in Section~\ref{sec:RegBayes}, when we introduce additional constraints, the
post-data posterior $q(\Thetav)$ is different from the Bayesian posterior
$p(\Thetav|\data)$, and moreover, it could even not be obtainable by the
conventional Bayesian inference via Bayes' rule}.
In the sequel, in order to distinguish $q(\cdot)$ from the Bayesian posterior,
we will call it post-data posterior. The optimization formulation
in~(\ref{eq:varBayes}) implies that Bayes' rule is an information projection
procedure that projects a prior density to a post-data posterior by taking
account of the observed data. In general, Bayes's rule is a special case of the
principle of minimum information~\cite{Williams:BayesCond1980}.

\subsection{Bayesian Methods in Machine Learning}

Bayesian statistics has been applied to almost every ML task, ranging from the
single-variate regression/classification to the structured output predictions
and to the unsupervised/semi-supervised learning scenarios~\cite{Bishop:PRML}.
In essence, however, there are several basic tasks that we briefly review below.

{\bf Prediction}: After training, Bayesian models make predictions using the distribution:
\begin{eqnarray}\label{eq:prediction}
p(\xv | \data ) = \! \int p(\xv, \Thetav | \data) \ud \Thetav 
= \! \int p(\xv| \Thetav , \data) p(\Thetav | \data) \ud \Thetav,
\end{eqnarray}
where $p(\xv | \Thetav, \data)$ is often simplified as $p(\xv | \Thetav)$ due to
the i.i.d assumption of the data when the model is given. Since the integral is
taken over the posterior distribution, the training data is considered.

{\bf Model Selection}: Model selection is a fundamental problem in statistics
and machine learning~\cite{Kadane:2004}.
Let $\model$ be a family of models, where each model is indexed by a set of
parameters $\Thetav$. Then, the marginal likelihood of the model family (or
model evidence) is
\begin{eqnarray}\label{eq:marginal}
p(\data | \model) = \int p(\data | \Thetav) p(\Thetav | \model) \ud \Thetav,
\end{eqnarray}
where $p(\Thetav | \model)$ is often assumed to be uniform if no strong prior
exists.

For two different model families $\model_1$ and $\model_2$, the ratio of model
evidences $\kappa = \frac{p(\data | \model_1)}{p(\data | \model_2)}$ is called
Bayes factor~\cite{Kass:1995}. The advantage of using Bayes factors for model
selection is that it automatically and naturally includes a penalty for
including too much model structure~\cite[Chap 3]{Bishop:PRML}. Thus, it guards
against overfitting. For models where an explicit version of the likelihood is
not available or too costly to evaluate, approximate Bayesian computation (ABC)
can be used for model selection in a Bayesian
framework~\cite{Grelaud:2009,Turnera:2012}, while with the caveat that
approximate-Bayesian estimates of Bayes factors are often
biased~\cite{Robert:2011}.


\subsection{Approximate Bayesian Inference}

Though conceptually simple, Bayesian inference has computational difficulties, which arise from the
intractability of high-dimensional integrals as involved in the posterior and in
Eq.s~(\ref{eq:prediction}, \ref{eq:marginal}). These are typically not only
analytically intractable but also difficult to obtain numerically.
Common practice resorts to approximate methods, which can be grouped into two
categories\footnote{Both maximum likelihood estimation (MLE),
$\hat{\Thetav}_{\textrm{MLE}} = \argmax_{\Thetav} p(\data | \Thetav)$, and
maximum a posterior estimation (MAP), $\hat{\Thetav}_{\textrm{MAP}} =
\argmax_{\Thetav} p_0(\Thetav) p(\data | \Thetav)$, can be seen as the third
type of approximation methods to do Bayesian inference. We omit them since they
examine only a single point,
and so can neglect the potentially large distributions in the integrals.} --- variational methods and Monte Carlo methods.

\subsubsection{Variational Bayesian Methods}\label{sec:variational}

Variational methods have a long history in physics, statistics, control theory
and economics. In machine learning, variational formulations appear naturally in
regularization theory, maximum entropy estimates, and approximate inference in
graphical models.
We refer the readers to the seminal book~\cite{Wainwright:2008} and the nice
short overview~\cite{Jordan:1999} for more details. A variational method
basically consists of two parts:
\begin{enumerate}
\item cast the problems as some optimization problems; 
\item find an approximate solution when the exact solution is not feasible.
\end{enumerate}
For Bayes' rule, we have provided a variational formulation
in~(\ref{eq:varBayes}), which is equivalent to minimizing the KL-divergence
between the variational distribution $q(\Thetav)$ and the target posterior
$p(\Thetav | \data)$.
We can also show that the negative of the objective in~(\ref{eq:varBayes}) is a lower bound of the evidence (i.e., log-likelihood): 
\begin{eqnarray}\label{eq:ELBO}
\log p(\data) \geq \ep_q[ \log p(\Thetav, \data) ] - \ep_q[ \log q(\Thetav) ].
\end{eqnarray}
Then, variational Bayesian methods maximize the Evidence Lower BOund (ELBO):
\begin{eqnarray}
\max_{q \in \prob} \ep_q[ \log p(\Thetav, \data) ] - \ep_q[ \log q(\Thetav) ],
\end{eqnarray}
whose solution is the target posterior if no assumptions 
are made.

However, in many cases it is intractable to calculate the target posterior.
Therefore, to simplify the optimization, the variational distribution is often
assumed to be in some parametric family, e.g., $q_{\phiv}(\Thetav)$, and has
some mean-field representation
\begin{eqnarray}
q_{\phiv}(\Thetav) = \prod_{i} q_{\phi_i}(\Thetav_i),
\end{eqnarray}
where $\{ \Thetav_i \}$ represent a partition of $\Thetav$.  Then, the problem
transforms to find the best parameters $\hat{\phiv}$ that maximize the ELBO,
which can be solved with numerical optimization methods. For example, with the
factorization assumption, coordinate descent is often used to iteratively solve
for $\phi_i$ until reaching some local optimum. Once a variational approximation
$q^\ast$ is found, the Bayesian integrals can be approximated by replacing
$p(\Thetav | \data)$ by $q^\ast$. In many cases, the model $\Thetav$ consists of
parameters $\thetav$ and hidden variables $\hv$. Then, if we make the
(structured) mean-field assumption
that $q(\thetav, \hv) = q(\thetav) q(\hv)$, the variational problem can be
solved by a variational Bayesian EM algorithm~\cite{Beal:2003}, which
alternately updates $q(\hv)$ at the variational Bayesian E-step and updates
$q(\thetav)$ at the variational Bayesian M-step.

\subsubsection{Monte Carlo Methods}

Monte Carlo (MC) methods represent a diverse class of algorithms that rely on
repeated random sampling to compute the solution to problems whose solution
space is too large to explore systematically or whose systemic behavior is too
complex to model. The basic idea of MC methods is to draw a set of i.i.d samples
$\{ \Thetav_i \}_{i=1}^N$ from a target distribution $p(\Thetav)$ and use the
empirical distribution $\hat{p}(\cdot) = \frac{1}{N} \sum_{i=1}^N
\delta_{\Thetav_i}(\cdot),$ to approximate the target distribution, where
$\delta_{\Thetav_i}(\cdot)$ is the delta-Dirac mass located at $\Thetav_i$.
Consider the common operation on calculating the expectation of some function
$\phi$ with respect to a given distribution. Let $p(\Thetav) = \bar{p}(\Thetav)
/ Z$ be the density of a probability distribution, where
$\bar{p}(\Thetav)$ is the unnormalized version that can be computed pointwise up to a normalizing constant $Z$.
The expectation of interest is
\begin{eqnarray}
I = \int \phi(\Thetav) p(\Thetav) \ud \Thetav.
\end{eqnarray}
Replacing $p(\cdot)$ by $\hat{p}(\cdot)$, we get the unbiased Monte Carlo estimate of this quantity: 
\begin{eqnarray}
\hat{I}_{\textrm{MC}} = \frac{1}{N} \sum_{i=1}^N \phi(\Thetav_i).
\end{eqnarray}

Asymptotically, when $N \rightarrow \infty$ the estimate $\hat{I}_{\textrm{MC}}$ 
will almost surely converge to $I$ by the strong law of large numbers. In
practice, however, we often cannot sample from $p$ directly. Many methods have
been developed, such as rejection sampling and importance sampling, which
however often suffer from severe limitations in high dimensional spaces. We
refer the readers to the book~\cite{Robert:2005} and the review
article~\cite{Andrieu:2003} for details. Below, we introduce Markov chain Monte
Carlo (MCMC), a very general and powerful framework that allows sampling from a
broad family of distributions and scales well with the dimensionality of the
sample space. More importantly, many advances have been made on scalable MCMC
methods for Big Data, which will be discussed later.

An MCMC method constructs an ergodic $p$-stationary Markov chain sequentially.
Once the chain has converged (i.e., finishing the burn-in phase), we can use the
samples to estimate $I$. The Metropolis-Hastings
algorithm~\cite{Metropolis:1953,Hastings:1970} constructs such a chain by using
the following rule to transit from the current state $\Thetav_t$ to the next
state $\Thetav_{t+1}$:
\begin{enumerate}
\item draw a candidate state $\Thetav^\prime$ from a proposal distribution $q(\Thetav | \Thetav_t)$;
\item compute the acceptance probability:
\begin{eqnarray}\label{eq:MH}
A(\Thetav^\prime, \Thetav_t) \triangleq \min\left(1,  \frac{\bar{p}(\Thetav^\prime ) q(\Thetav_t | \Thetav^\prime) }{\bar{p}(\Thetav_t) q(\Thetav^\prime | \Thetav_t) } \right).
\end{eqnarray}
\item draw $\gamma \sim \textrm{Uniform}[0,1]$. If $\gamma <  A(\Thetav^\prime, \Thetav_t)$ set $\Thetav_{t+1} \gets \Thetav^\prime$, otherwise set $\Thetav_{t+1} \gets \Thetav_t$.
\end{enumerate}
Note that for Bayesian models, each MCMC step involves an evaluation of the full
likelihood to get the (unnormalized) posterior $\bar{p}(\Thetav)$, which can be
prohibitive for big learning with massive data sets. We will revisit this
problem later.

One special type of MCMC methods is the Gibbs sampling~\cite{Geman:1984}, which
iteratively draws samples from local conditionals. Let $\Thetav$ be a
$M$-dimensional vector. The standard Gibbs sampler performs the following steps
to get a new sample $\Thetav^{(t+1)}$:
\begin{enumerate}
\item draw a sample $\theta_1^{(t+1)} \sim p(\theta_1 | \theta_2^{(t)}, \cdots, \theta_M^{(t)} )$;
\item for $j=2:M-1$, draw a sample $$\theta_j^{(t+1)} \sim p(\theta_j | \theta_1^{(t+1)}, \cdots, \theta_{j-1}^{(t+1)}, \theta_{j+1}^{t}\cdots, \theta_M^{t} );$$
\item draw a sample $\theta_M^{(t+1)} \sim p(\theta_M | \theta_1^{(t+1)}, \cdots, \theta_{M-1}^{(t+1)})$.
\end{enumerate}

One issue with MCMC methods is that the convergence rate can be prohibitively
slow even for conventional applications. Extensive efforts have been spent to
improve the convergence rates. For example, hybrid Monte Carlo methods explore
gradient information to improve the mixing rates when the model parameters are
continuous, with representative examples of Langevin dynamics and Hamiltonian
dynamics~\cite{Neal:10}. Other improvements include population-based MCMC
methods~\cite{Jasra:2007} and annealing methods~\cite{Geyer:1995} that can
sometimes handle distributions with
multiple modes. Another useful technique to develop simpler or more efficient MCMC methods is data augmentation~\cite{Tanner:1987,DykMeng2001,Neal:slice2003}, which introduces auxiliary variables to transform marginal dependency into a set of conditional independencies.
For Gibbs samplers, blockwise Gibbs sampling and partially collapsed Gibbs (PCG)
sampling~\cite{pcgs} often improve the convergence. A PCG sampler is as simple as an ordinary Gibbs sampler,
but often improves the convergence by replacing some of the conditional
distributions of an ordinary Gibbs sampler with conditional distributions of
some marginal distributions.

\subsection{FAQ} \label{sec:FAQ}

Common questions regarding Bayesian methods are: 

{\bf Q: Why should I use Bayesian methods?}

{\bf A:} There are many reasons for choosing Bayesian methods, as discussed in the Introduction. 
A formal theoretical argument is provided by the classic de Finitti theorem, which states that:
If $(\xv_1, \xv_2, \dots)$ are infinitely exchangeable, then for any $N$
\begin{eqnarray}
p(\xv_1, \dots, \xv_N) = \int \left( \prod_{i=1}^N p(\xv_i | \thetav) \right) \ud P(\thetav)
\end{eqnarray}
for some random variable $\thetav$ and probability measure $P$.
The infinite exchangeability is an often satisfied property. For example, any
i.i.d data are infinitely exchangeable. Moreover, the data whose ordering
information is not informative is also infinitely exchangeable, e.g., the
commonly used bag-of-words representation of documents~\cite{Blei:03} and
images~\cite{Fei-Fei:05}.

{\bf Q: How should I choose the prior?}

{\bf A:} There are two schools of thought, namely, objective Bayes and
subjective Bayes. For objective Bayes, an improper noninformative prior (e.g.,
the Jeffreys prior~\cite{Jeffreys:1945} and the maximum-entropy
prior~\cite{Jaynes:1968}) is used to capture ignorance, which admits good
frequentist properties. In contrast, subjective Bayesian methods embrace the
influence of priors.
A prior may have some parameters $\lambda$. Since it is often difficult to
elicit an honest prior, e.g., setting the true value of $\lambda$, two practical
methods are often used. One is hierarchical Bayesian methods, which assume a
hyper-prior on $\lambda$ and define the prior as a marginal distribution:
\begin{eqnarray}
p_0(\Thetav ) = \int p_0(\Thetav | \lambda) p(\lambda) \ud \lambda.
\end{eqnarray}
Though $p(\lambda)$ may have hyper-parameters as well, it is commonly believed
that these parameters will have a weak influence as long as they are far from
the likelihood model, thus can be fixed at some convenient values or put another
layer of hyper-prior.

Another method is {\it empirical Bayes}, which adopts a data-driven estimate
$\hat{\lambda}$ and uses $p_0(\Thetav | \hat{\lambda})$ as the prior. Empirical
Bayes can be seen as an approximation to the hierarchical approach, where
$p(\lambda)$ is approximated by a delta-Dirac mass
$\delta_{\hat{\lambda}}(\lambda)$. One common choice is maximum marginal
likelihood estimate, that is, $\hat{\lambda} = \argmax_{\lambda} p(\data |
\lambda)$. Empirical Bayes has been applied in many problems, including variable
section~\cite{George:2000} and nonparametric Bayesian
methods~\cite{McAuliffe:2006}.
Recent progress has been made on characterizing the conditions when empirical
Bayes merges with the Bayesian inference~\cite{Petrone:2014} as well as the
convergence rates of empirical Bayes methods~\cite{Donnet:2014}.

In practice, another important consideration is the tradeoff between model
capacity and computational cost. If a prior is conjugate to the likelihood, the
posterior inference will be relatively simpler in terms of computation and
memory demands, as the posterior belongs to the same family as the prior.

{\bf Example 1: Dirichlet-Multinomial Conjugate Pair}
Let $\xv \in \{0,1\}^V$ be a one-hot representation of a discrete variable with $V$ possible values. 
It is easy to verify that for the multinomial likelihood, $p(\xv | \thetav) =
\prod_{k=1}^V \theta_k^{x_k}$, the conjugate prior is a Dirichlet distribution,
$p_0(\thetav | \alphav) = \textrm{Dir}(\alphav) = \frac{1}{Z} \prod_{k=1}^V
\theta_k^{\alpha_k - 1}$, where $\alphav$ is the hyper-parameter and $Z$ is the
normalization factor. In fact, the posterior distribution is
$\textrm{Dir}(\alphav + \xv)$.

\begin{figure}[t]\vspace{-.2cm}
\centering
{\hfill
\subfigure[]{\includegraphics[height = .34\textwidth]{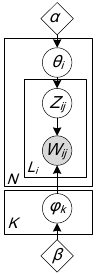}}\hfill
\subfigure[]{\includegraphics[height = .34\textwidth]{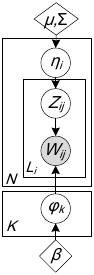}}\hfill
\subfigure[]{\includegraphics[height = .34\textwidth]{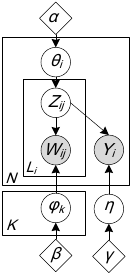}}\hfill}\vspace{-.2cm}
\caption{Graphical models of (a) LDA~\cite{Blei:03}; (b) logistic-normal topic model~\cite{Blei:06}; and (c) supervised LDA.   } 
\label{fig:LDA}\vspace{-.3cm}
\end{figure}

A popular Bayesian model that explores such conjugacy is latent Dirichlet
allocation~(LDA)~\cite{Blei:03}, as illustrated in Fig.~\ref{fig:LDA}(a).\footnote{All the figures are drawn by the authors with full copyright.} LDA
posits that each document $\wv_i$ is an admixture of a set of $K$ topics, of
which each topic $\psiv_k$ is a unigram distribution over a given vocabulary.
The generative process is as follows:
\begin{enumerate}
\item draw $K$ topics $\psiv_k \sim \textrm{Dir}(\betav)$
\item for each document $i \in [N]$:
\begin{enumerate}
\item draw a topic mixing vector $\thetav_i \sim \textrm{Dir}(\alphav)$
\item for each word $j \in [L_i]$ in document $i$:
\begin{enumerate}
\item draw a topic assignment $z_{ij} \sim \textrm{Multi}(\thetav_i)$
\item draw a word $w_{ij} \sim \textrm{Multi}(\psiv_{z_{ij}})$.
\end{enumerate}
\end{enumerate}
\end{enumerate}
LDA has been popular in many applications. However, a conjugate prior can be
restrictive. For example, the Dirichlet distribution does not impose correlation
between different parameters, except the normalization constraint. In order to
obtain more flexible models, a non-conjugate prior can be chosen.

{\bf Example 2: Logistic-Normal Prior}
A logistic-normal distribution~\cite{Aitchison:80} provides one way to impose correlation structure among the multiple dimensions of $\thetav$. It is defined as follows:
\begin{eqnarray}
\etav \sim \mathcal{N}(\muv, \Sigma),~ \theta_k = \frac{e^{\eta_k}}{ \sum_j e^{\eta_j}}.
\end{eqnarray}
This prior has been used to develop correlated topic models (or logistic-normal
topic models)~\cite{Blei:06}, which can infer the correlation structure among
topics. However, the flexibility pays cost on computation, needing scalable
algorithms to learn large topic graphs~\cite{Chen:2013}.

\vspace{-.1cm}
\section{Big Bayesian Learning}
\vspace{-.1cm}

Though much more emphasis in big Bayesian learning has been put on scalable
algorithms and systems, substantial advances have been made on {\it adaptive}
and {\it flexible} Bayesian methods. This section reviews nonparametric Bayesian
methods for adaptively inferring model complexity and regularized Bayesian
inference for improving the flexibility via posterior regularization, while
leaving the large part of scalable algorithms and systems to next sections.

\vspace{-.1cm}
\subsection{Nonparametric Bayesian Methods}\label{sec:NPB}
\vspace{-.05cm}

For parametric Bayesian models, the parameter space is pre-specified. No matter
how the data changes, the number of parameters is fixed. This restriction may
cause limitations on model capacity, especially for big data applications, where
it may be difficult or even counter-productive to fix the number of parameters a
priori. For example, a Gaussian mixture model with a fixed number of clusters
may fit the given data set well; however, it may be sub-optimal to use the same
number of clusters if more data comes under a slightly changed distribution. It
would be ideal if the clustering models can figure out the unknown number of
clusters automatically. Similar requirements on automatical model selection
exist in feature representation learning~\cite{Bengio:2012} or factor analysis,
where we would like the models to automatically figure out the dimension of
latent features (or factors) and maybe also the topological structure among
features (or factors) at different abstraction levels~\cite{Adams:2011}.

Nonparametric Bayesian (NPB) methods provide an elegant solution to such needs
on automatic adaptation of model capacity when learning a single model. Such
adaptivity is obtained by defining stochastic processes on rich measure spaces.
Classical examples include Dirichlet process (DP), Indian buffet process (IBP),
and Gaussian process (GP). Below, we briefly review DP and IBP. We refer the
readers to the articles~\cite{Ghahramani:2013,Gershmana:2012,Muller:2004} for a
nice overview and the textbook~\cite{Hjort:2010} for a comprehensive treatment.

\subsubsection{Dirichlet Process}
\begin{figure}[t]\vspace{-.2cm}
\centering
\includegraphics[width = .45\textwidth]{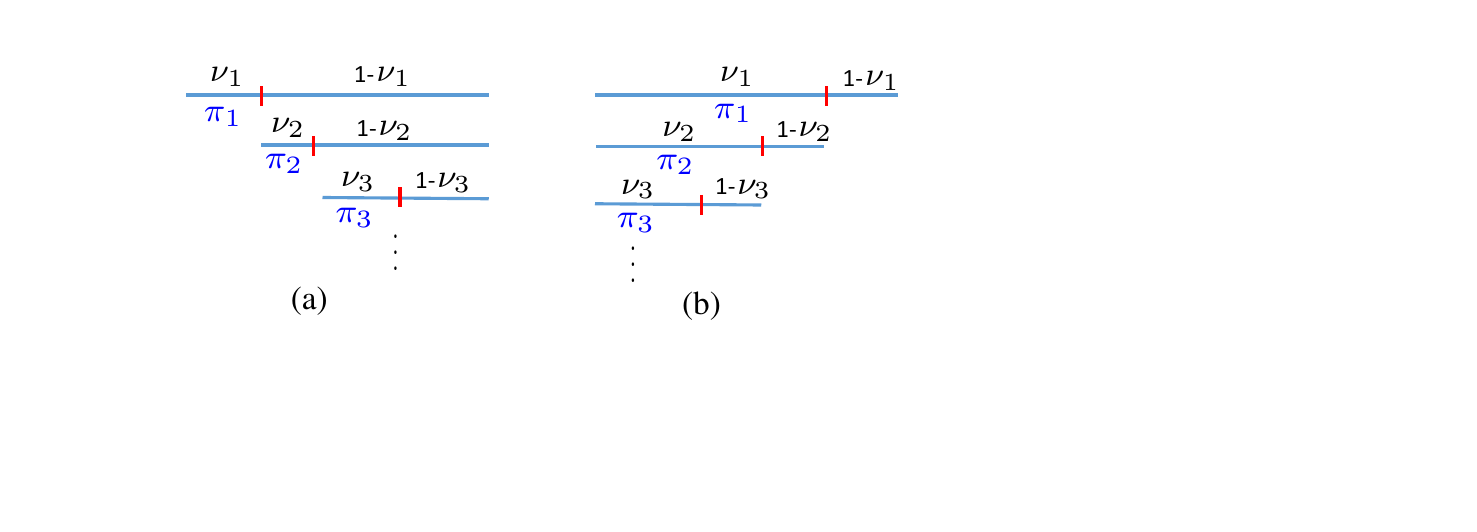}\vspace{-.2cm}
\caption{The stick-breaking process for: (a) DP; (b) IBP.} \label{fig:stickbreaking}\vspace{-.4cm}
\end{figure}

A DP defines the distribution of random measures. It was first developed
in~\cite{Ferguson:73}. Specifically, a DP is parameterized by a concentration
parameter $\alpha > 0$ and a base distribution $G_0$ over a measure space
$\Omega$. A random variable drawn from a DP, $G \sim \mathcal{DP}(\alpha, G_0)$,
is itself a distribution over $\Omega$. It was shown that the random
distributions drawn from a DP are discrete almost surely, that is, they place
the probability mass on a countably infinite collection of atoms, i.e.,
\begin{eqnarray}\label{eq:stick-breaking-DP}
G = \sum_{k=1}^\infty \pi_k \delta_{\theta_k},
\end{eqnarray}
where $\theta_k$ is the value (or location) of the $k$th atom independently
drawn from the base distribution $G_0$ and $\pi_k$ is the probability assigned
to the $k$th atom. Sethuraman~\cite{Sethuraman:94} provided a constructive
definition of $\pi_k$ based on a stick-breaking process as illustrated in
Fig.~\ref{fig:stickbreaking}(a).
Consider a stick with unit length. We break the stick into an infinite number of segments $\pi_k$ by the following process with $\nu_k \sim \textrm{Beta}(1, \alpha)$:
\begin{eqnarray}
\pi_1 = \nu_1, ~~\pi_k = \nu_k \prod_{j=1}^{k-1} (1- \nu_j),~k=2,3,\dots,\infty.
\end{eqnarray}
That is, we first choose a beta variable $\nu_1$ and break $\nu_1$ of the stick. Then, for the remaining segment, we draw another beta variable and break off that proportion of the remainder of the stick. Such a representation of DP 
provides insights for developing variational approximate inference algorithms~\cite{Blei:hdp06}.

DP is closely related to the Chinese restaurant process
(CRP)~\cite{Pitman:2002}, which defines a distribution over infinite partitions
of integers. CRP derives its name from a metaphor: Image a restaurant with an
infinite number of tables and a sequence of customers entering the restaurant
and sitting down. The first customer sits at the first table. For each of the
subsequent customers, she sits at each of the occupied tables with a probability
proportional to the number of previous customers sitting there, and at the next
unoccupied table with a probability proportional to $\alpha$. In this process,
the assignment of customers to tables defines a random partition.
In fact, if we repeatedly draw a set of samples from $G$, that is, $\thetav_i \sim G,~i \in [N]$, then it was shown that the joint distribution of $\thetav_{1:N}$
\begin{eqnarray}
p(\thetav_1, \dots, \thetav_N | \alpha, G_0) = \int \left( \prod_{i=1}^N p(\thetav_i | G) \right) \ud P(G | \alpha, G_0) \nonumber
\end{eqnarray}
exists a clustering property, that is, the $\thetav_i$s will share repeated
values with a non-zero probability. These shared values define a partition of
the integers from 1 to $N$, and the distribution of this partition is a CRP with
parameter $\alpha$. Therefore, DP is the de Finetti mixing distribution of CRP.

Antoniak~\cite{Antoniak:74} first developed DP mixture models by adding a data
generating step, that is, $\xv_i \sim p( \xv | \thetav_i),~i \in [N].$ Again,
marginalizing out the random distribution $G$, the DP mixture reduces to a CRP
mixture, which enjoys nice Gibbs sampling algorithms~\cite{Neal:2000}. For DP
mixtures, a slice sampler~\cite{Neal:slice2003} has been
developed~\cite{Walker:2007}, which transforms the infinite sum in
Eq.~(\ref{eq:stick-breaking-DP}) into a finite sum conditioned on some uniformly
distributed auxiliary variable.

\subsubsection{Indian Buffet Process}
A mixture model assumes that each data is assigned to one single component.
Latent factor models weaken this assumption by associating each data with some
or all of the components. When the number of components is smaller than the
feature dimension, latent factor models provide dimensionality reduction.
Popular examples include factor analysis, principal component analysis and
independent component analysis. The general assumption of a latent factor model
is that the observed data $\xv \in \mathbb{R}^P$ is generated by a noisy
weighted combination of latent factors, that is,
\begin{eqnarray}
\xv_i = W \zv_i + \epsilon_i,
\end{eqnarray}
where $W$ is a $P \times K$ factor loading matrix, with element $W_{mk}$
expressing how latent factor $k$ influences the observation dimension $m$;
$\zv_i$ is a $K$-dimensional vector expressing the activity of each factor; and
$\epsilon_i$ is a vector of independent noise terms (usually Gassian noise). In
the above models, the number of factors $K$ is assumed to be known. Indian
buffet process (IBP)~\cite{Griffiths:nips06} provides a nonparametric Bayesian
variant of latent factor models and it allows the number of factors to grow as
more data are observed.

Consider binary factors for simplicity\footnote{Real-valued factors can be
easily considered by defining $\hv_i = \zv_i \odot \muv_i$, where the binary
$\zv_i$ are $0/1$ masks to indicate whether a factor is active or not, and
$\muv_i$ are the values of the factors. }. Putting the latent factors of $N$
data points in a matrix $Z$, of which the $i$th row is $\zv_i$. IBP defines a
process over the space of binary matrixes with an unbounded number of columns.
IBP derives its name from a similar metaphor as CRP. Image a buffet with an
infinite number of dishes (factors) arranged in a line and a sequence of
customers choosing the dishes. Let $z_{ik}$ denote whether customer $i$ chooses
dish $k$. Then,
the first customer chooses $K_1$ dishes, where $K_1 \sim
\textrm{Poisson}(\alpha)$; and the subsequent customer $n~ (> 1)$ chooses:
\begin{enumerate}
\item each of the previously sampled dishes with probability $m_k / n$, where $m_k$ is the number of customers who have chosen dish $k$;
\item $K_i$ additional dishes, where $K_i \sim \textrm{Poisson}(\alpha / n)$.
\end{enumerate}

IBP plays the same role for latent factor models that CRP plays for mixture
models, allowing an unbounded number of latent factors. Analogous to the role
that DP is the de Finetti mixing distribution of CRP, the de Finetti mixing
distribution underlying IBP is a Beta process~\cite{Thibaux:2007}. IBP also
admits a stick-breaking representation~\cite{Teh:2007} as shown in
Fig.~\ref{fig:stickbreaking}(b), where the stick lengths are defined as
\begin{eqnarray}
\nu_k \sim \textrm{Beta}(\alpha, 1),~~ \pi_k = \prod_{j=1}^k \nu_j,~~ k=1,2,\dots, \infty.
\end{eqnarray}
Note that unlike the stick-breaking representation of DP, where the stick
lengths sum to 1, the stick lengths here need not sum to 1. Such a
representation has lead to the developments of Monte Carlo~\cite{Teh:2007} as
well as variational approximation inference algorithms~\cite{Doshi-Velez:2009}.

\iftrue
\subsubsection{Gaussian Process}

Kernel machines (e.g., support vector machines)~\cite{Hofmann:2008} represent an important class of methods in machine learning and has received extensive attention. Gaussian processes (GPs) provide a principled, practical, probabilistic approach to learning in kernel machines. A Gaussian process is defined on the space of continuous functions~\cite{Rasmussen:2006}. In machine learning, the prime use of GPs is to learn the unknown mapping function from inputs to outputs for supervised learning.

Take the simple linear regression model as an example. Let $\xv \in \mathbb{R}^M$ be an input data point and $y \in \mathbb{R}$ be the output. A linear regression model is $$f(\xv) = \thetav^\top \phi(\xv),~~ y = f(\xv) + \epsilon,$$ where $\phi(\xv)$ is a vector of features extracted from $\xv$, and $\epsilon$ is an independent noise. For the Gaussian noise, e.g., $\epsilon \sim \mathcal{N}(0, \sigma^2 I)$, the likelihood of $y$ conditioned on $\xv$ is also a Gaussian, that is, $p(y | \xv, \thetav) = \mathcal{N}(f(\xv), \sigma^2 I)$. Consider a Bayesian approach, where we put a zero-mean Gaussian prior, $\thetav \sim \mathcal{N}(0, \Sigma)$. Given a set of training observations $\data = \{(\xv_i, y_i) \}_{i=1}^N$. Let $X$ be the $M \times N$ design matrix, and $\yv$ be the vector of the targets. By Bayes' theorem, we can easily derive that the posterior is also a Gaussian distribution (see~\cite{Rasmussen:2006} for more details)
\begin{eqnarray}
p(\thetav | X, \yv) = \mathcal{N}\left( \frac{1}{\sigma^2} A^{-1} \Phi \yv, A^{-1} \right),
\end{eqnarray}
where $A^{-1} = \sigma^{-2} \Phi \Phi^\top + \Sigma^{-1}$ and $\Phi = \phi(X)$. For a test example $\xv_\ast$, we can also derive that the distribution of the predictive value $f_\ast \triangleq f(\xv_\ast)$ is also a Gaussian:
\begin{eqnarray}
p(f_\ast | \xv_\ast, X, \yv) = \mathcal{N}\left( \frac{1}{\sigma^2} \phi_\ast^\top A^{-1} \Phi \yv,  \phi_\ast^\top A^{-1} \phi_\ast \right),
\end{eqnarray}
where $\phi_\ast \triangleq \phi(\xv_\ast)$. In some equivalent form, the Gaussian mean and covariance only involve the inner products in input space. Therefore, the kernel trick can be explored in such models, which avoids the explicit evaluation of the feature vectors.

The above Bayesian linear regression model is a very simple example of Gaussian processes. In the most general form, Gaussian processes define a stochastic process over functions $f(\xv)$. A GP is characterized by a mean function $m(\xv)$ and a covariance function $\kappa(\xv, \xv^\prime)$, denoted by $f(\xv) \sim \mathcal{GP}(m(\xv), \kappa(\xv, \xv^\prime)$. Given any finite set of observations $\xv_1, \dots, \xv_n$, the function values\footnote{The function values are random variables due to the randomness of $f$.} $(f(\xv_1), \dots, f(\xv_n))$ follow a multivariate Gaussian distribution with mean $(m(\xv_1), \dots, m(\xv_n))$ and covariance $K:~K(i,j) = \kappa(\xv_i, \xv_j)$. The above definition with any finite collection of function values guarantee to define a stochastic process (i.e., Gaussian process), by examining the consistency requirement of the Kolmogorov extension theorem.

Gaussian processes have also been used in classification tasks, where the likelihood is often non-conjugate to the Gaussian process prior, therefore requiring approximate inference algorithms, including both variational and Monte Carlo methods. Other research has considered Gaussian process latent variable models (GP-LVM)~\cite{Lawrence:2005}.
\fi

\subsubsection{Extensions}
To meet the flexibility and adaptivity requirements of big learning, many recent
advances have been made on developing sophisticated NPB methods for modeling
various types of data, including grouped data, spatial data, time series, and
networks.

Hierarchical models are natural tools to describe grouped data, e.g., documents
from different source domains. Hierarchical Dirichlet process
(HDP)~\cite{Teh:2006} and hierarchical Beta process~\cite{Thibaux:2007} have
been developed, allowing an infinite number of latent components to be shared by
multiple domains. The work~\cite{Adams:2011} presents a cascading IBP (CIBP) to
learn the topological structure of multiple layers of latent features, including
the number of layers, the number of hidden units at each layer, the connection
structure between units at neighboring layers, and the activation function of
hidden units. The recent work~\cite{Dallaire:2014} presents an extended CIBP
process to generate connections between non-consecutive layers.

Another dimension of the extensions concerns modeling the dependencies between
observations in a time series. For example, DP has been used to develop the
infinite hidden Markov models~\cite{Beal:iHMM07}, which posit the same
sequential structure as in the hidden Markov models, but allowing an infinite
number of latent classes. In~\cite{Teh:2006}, it was shown that iHMM is a
special case of HDP. The recent work~\cite{Zhang:2014} presents a max-margin
training of iHMMs under the regularized Bayesian framework, as will be reviewed
shortly.

Finally, for spatial data, modeling dependency between nearby data points is
important. Recent extensions of Bayesian nonparametric methods include the
dependent Dirichlet process~\cite{MacEachern:1999}, spatial Dirichlet
process~\cite{Duan:2007}, distance dependent CRP~\cite{Blei:2010}, dependent
IBP~\cite{Williamson:2010}, and distance dependent IBP~\cite{Gershman:2011}. For
network data analysis (e.g., social networks, biological networks, and citation
networks), recent extensions include the nonparametric Bayesian relational
latent feature models for link prediction~\cite{Miller:nips09,Zhu:ICML12}, which
adopt IBP to allow for an unbounded number of latent features, and the
nonparametric mixed membership stochastic block models for community
discovery~\cite{Gopalan:2013,Kim:2013}, which use HDP to allow mixed membership
in an unbounded number of latent communities.

\begin{figure*}[t]\vspace{-.2cm}
\centering
{\hfill
\subfigure[]{\includegraphics[height = .15\textwidth]{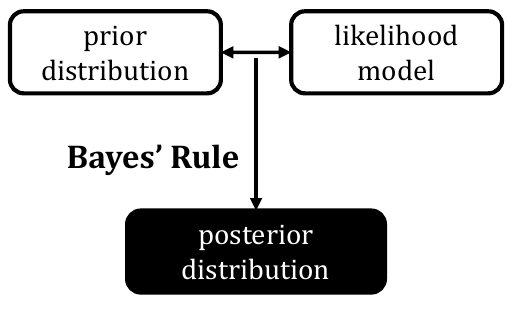}}\hfill
\subfigure[]{\includegraphics[height = .15\textwidth]{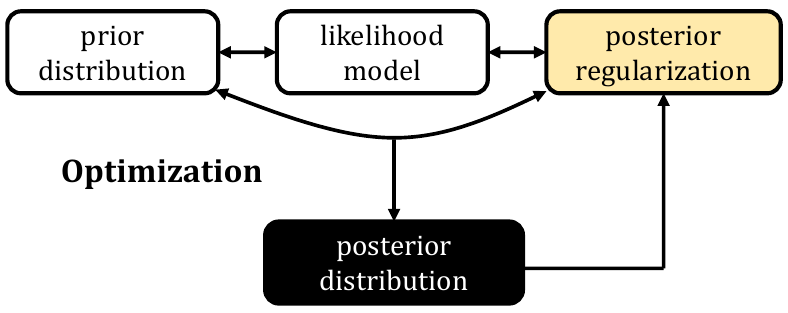}}\hfill}\vspace{-.2cm}
\caption{(a) Bayesian inference with the Bayes' rule; and (b) regularized Bayesian inference (RegBayes) which solves an optimization problem with a posterior regularization term to incorporate rich side information.}\label{fig:RegBayes}\vspace{-.4cm}
\end{figure*}

\subsection{Regularized Bayesian Inference}\label{sec:RegBayes}

Regularized Bayesian inference (RegBayes)~\cite{Zhu:regBayes14} represents one
recent advance that extends the scope of Bayesian methods on incorporating rich
side information. Recall that the classic Bayes' theorem is equivalent to a
variational optimization problem as in~(\ref{eq:varBayes}). RegBayes builds on
this formulation and defines the generic optimization problem
\begin{eqnarray}\label{eq:regBayes}
\min_{q(\Thetav) \in \prob} \KL(q(\Thetav) \Vert p(\Thetav | \data) ) + c \cdot \Omega(q(\Thetav); \data),
\end{eqnarray}
where $\Omega(q(\Thetav); \data)$ is the {\it posterior regularization} term;
$c$ is a nonnegative regularization parameter; and $p(\Thetav | \data)$ is the
ordinary Bayesian posterior.
Fig.~\ref{fig:RegBayes} provides a high-level comparison between RegBayes and
Bayes' rule. 
Several questions need to be answered in order to solve practical
problems.

{\bf Q: How to define the posterior regularization?}

{\bf A:} In general, posterior regularization can be any informative constraints
that are expected to regularize the properties of the posterior distribution. It
can be defined as the large-margin constraints to enforce a good prediction
accuracy~\cite{Zhu:jmlr12}, or the logic constraints to incorporate expert
knowledge~\cite{Mei:robustBayes14}, or the sparsity
constraints~\cite{Koyejo:uai13}.

{\bf Example 3: Max-margin LDA} Following the paradigm of ordinary Bayes, a
supervised topic model is often defined by augmenting the likelihood model. For
example, the supervised LDA (sLDA)~\cite{Blei:07} has a similar structure as LDA
(see Fig.~\ref{fig:LDA}(c)), but with an additional likelihood $p(y_d | \zv_d,
\etav)$ to describe labels. Such a design can lead to an imbalanced combination
of the word likelihood $p(\wv_d | \zv_d, \psiv)$ and the label likelihood
because a document often has tens or hundreds of words while only one label. The
imbalance problem causes unsatisfactory prediction results~\cite{Zhu:ACL13}.

To improve the discriminative power of supervised topic models, the max-margin
MedLDA has been developed, under the RegBayes framework. Consider binary
classification for simplicity. In this case, we have $\Thetav = \{\thetav_i,
\zv_i, \psiv_k\}$. Let $f(\etav, \zv_i) = \etav^\top \bar{\zv}_i$ be the
discriminant function\footnote{We ignore the offset for simplicity.}, where
$\bar{\zv}_i$ is the average topic assignments, with $\bar{z}_i^k =
\frac{1}{L_i} \sum_j \indicator(z_{ij} = k)$.
The posterior regularization can be defined in two ways:

{\it Averaging classifier}:  An averaging classifier makes predictions using the
expected discriminant function, that is, $\hat{y}(q) = \textrm{sign}( \ep_q[
f(\etav, \zv) ] )$. Let $(x)_+ = \max(0, x)$. Then, the posterior regularization
$$\Omega^{\textrm{Avg}}(q(\Thetav); \data) = \sum_{i=1}^N ( 1 - y_i \ep_q[
f(\etav, \zv_i) ] )_{+}$$ is an upper bound of the training error, therefore a
good surrogate loss for learning. This strategy has
been adopted in MedLDA~\cite{Zhu:jmlr12}.

{\it Gibbs classifier}: A Gibbs classifier (or stochastic classifier) randomly
draws a sample $(\etav, \zv_d)$ from the target posterior $q(\Thetav)$ and makes
predictions using the latent prediction rule, that is, $\hat{y}(\etav, \zv_i) =
\textrm{sign} f(\etav, \zv_i)$. Then, the posterior regularization is defined as
$$\Omega^{\textrm{Gibbs}}(q(\Thetav); \data) = \ep_q\left[ \sum_{i=1}^N (1 - y_i
f(\etav, \zv_i) )_+ \right].$$
This strategy has been adopted to develop Gibbs MedLDA~\cite{Zhu:ICML13}.

The two strategies are closely related, e.g., we can show that
$\Omega^{\textrm{Gibbs}}(q(\Thetav)) $ is an upper bound of
$\Omega^{\textrm{Avg}}(q(\Thetav))$. The formulation with a Gibbs classifier can
lead to a scalable Gibbs sampler by using data augmentation
techniques~\cite{Zhu:KDD13}. If a logistic log-loss is adopted to define the
posterior regularization, an improved sLDA model can be developed to address the
imbalance issue and lead to significantly more accurate
predictions~\cite{Zhu:ACL13}.

{\bf Q: What is the relationship between prior, likelihood, and posterior regularization?}

{\bf A:} Though the three parts are closely connected, there are some key
differences. First, prior is chosen before seeing data, while both likelihood
and posterior regularization depend on the data. Second, different from the
likelihood, which is restricted to be a normalized distribution, no constraints
are imposed on the posterior regularization. Therefore, posterior regularization
is much more flexible than prior or likelihood. In fact, it can be shown that
(1) putting constraints on priors is a special case of posterior regularization,
where the regularization term does not depend on data; and (2) RegBayes can be
more flexible than standard Bayes' rule, that is, there exists some RegBayes
posterior distributions that are not achievable by the Bayes'
rule~\cite{Zhu:regBayes14}.

{\bf Q: How to solve the optimization problem?}

{\bf A:} The posterior regularization term affects the difficulty of solving
problem~(\ref{eq:regBayes}). When the regularization term is a convex functional
of $q(\Thetav)$, which is common in many applications such as the above
max-margin formulations, the optimal solution can be characterized in a general
from via convex duality theory~\cite{Zhu:regBayes14}. When the regularization
term is non-convex, a generalized representation theorem can also be derived,
but requires more effects on dealing with the non-convexity~\cite{Koyejo:uai13}.

\begin{figure*}[t]\vspace{-.2cm}
\centering
\includegraphics[width = .84\textwidth]{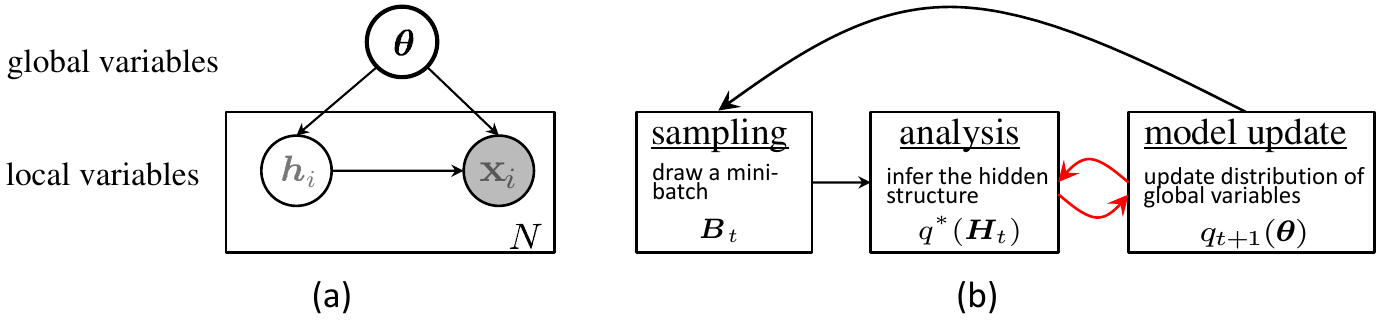}\vspace{-.3cm}
\caption{(a) The general structure of Bayesian latent variable models, where $\hv_i$ denotes the local latent variables for each data $i$; (b) the process of stochastic variational inference, where the red arrows denote that in practice we may need multiple iterations between ``analysis" and ``model update" to have fast convergence.}\label{fig:SVI}\vspace{-.4cm}
\end{figure*}

\section{Scalable Algorithms}

To deal with big data, the posterior inference algorithms should be scalable.
Significant advances have been made in two aspects: (1) using random sampling to
do stochastic or online Bayesian inference; and (2) using multi-core and
multi-machine architectures to do parallel and distributed Bayesian inference.

\subsection{Stochastic Algorithms}

In Big Learning, the intriguing results of~\cite{Bottou:2008} suggest that an
algorithm as simple as stochastic gradient descent (SGD) can be optimally
efficient in terms of ``number of bits learned per unit of computation". For
Bayesian models, both stochastic variational and stochastic Monte Carlo methods
have been developed to explore the redundancy of data relative to a model by
subsampling data examples for every update and reasoning about the uncertainty
created in this process~\cite{Welling:2014}. We overview each type in turn.

\subsubsection{Stochastic Variational Methods}

As we have stated in Section~\ref{sec:variational}, variational methods solve an
optimization problem to find the best approximate distribution to the target
posterior. When the variational distribution is characterized in some parametric
form, this problem can be solved with stochastic gradient descent (SGD)
methods~\cite{Boyd:04} or the adaptive SGD~\cite{Duchi:2011}. A SGD method
randomly draws a subset $B_t$ and updates the variational parameters using the
estimated gradients, that is,
\begin{eqnarray}
\phiv_{t+1} \gets \phiv_t + \epsilon_t \left( \nabla_{\phiv} \KL( q \Vert p_0(\thetav) ) - \nabla_{\phiv} \ep_q[ \log p(\data | \thetav) ] \right), \nonumber
\end{eqnarray}
where the full data gradient is approximated as
\begin{eqnarray}\label{eq:stochastic-var}
\nabla_{\phiv} \ep_q[ \log p(\data | \thetav) ] \approx  \frac{N}{|B_t|} \sum_{i \in B_t}  \nabla_{\phiv} \ep_q[ \log p(\xv_i | \thetav) ] ,
\end{eqnarray}
and $\epsilon_t$ is a learning rate. If the noisy gradient is an unbiased
estimate of the true gradient, the procedure is guaranteed to approach the
optimal solution when the learning rate is appropriately set~\cite{Bottou:1998}.

For Bayesian latent variable models, we need to infer the latent variables when
performing the updates. In general, we can group the latent variables into two
categories --- global variables and local variables. Global variables correspond
to the model parameters $\thetav$ (e.g., the topics $\psiv$ in LDA), while local
variables represent some hidden structures of the data (e.g., the topic
assignments $\zv$ in an LDA with the topic mixing proportions collapsed out).
Fig.~\ref{fig:SVI} provides an illustration of such models and the stochastic
variational inference, which consists of three steps:
\begin{enumerate}
\item randomly draw a mini-batch $B_t$ of data samples;
\item infer the local latent variables for each data in $B_t$;
\item update the global variables.
\end{enumerate}

However, the standard gradients over the parameters $\phiv$ may not be the most
informative direction (i.e., the steepest direction) to search for the
distribution $q$. A better way is to use natural gradient~\cite{Amari:1998},
which is the steepest search direction in a Riemannian manifold space of
probability distributions~\cite{Hoffman:2013}. To reduce the efforts on
hand-tuning the learning rate, which often influences the performance much, the
work~\cite{Ranganath:2013} presents an adaptive learning rate
while~\cite{Snoek:2012} adopts Bayesian optimization to search for good learning
rates, both leading to faster convergence. By borrowing the gradient averaging
ideas from stochastic optimization, \cite{Mandt:2014} proposes to use smoothed
gradients in stochastic variational inference to reduce the variance (by
trading-off the bias). Stochastic variational inference methods have been
studied for many Bayesian models, such as LDA and hierarchical
Dirichlet process~\cite{Hoffman:2013}.

In many cases, the ELBO and its gradient may be intractable to compute due to
the intractability of the expectation over variational distributions. Two types
of methods are commonly used to address this problem. First, another layer of
variational bound is derived by introducing additional variational parameters.
This has been used in many examples, such as the logistic-normal topic
models~\cite{Blei:06} and supervised LDA~\cite{Blei:07}. For such methods, it is
important to develop tight variational bounds for specific
models~\cite{Marlin:2011}, which is still an active area. Another type of
methods is to use Monte Carlo estimates of the variational bound as well as its
gradients. Recent work includes the stochastic approximation scheme with
variance reduction~\cite{Paisley:2012,Ranganath:2013} and the auto-encoding
variational Bayes (AEVB)~\cite{Kingma:2014} that learns a neural network (a.k.a
recognition model) to represent the variational distribution for continuous
latent variables.

Consider the model with one layer of continuous latent variables $\hv_i$ in
Fig.~\ref{fig:SVI} (a). Assume the variational distribution $q_{\phiv}(\Thetav)
= q_{\phiv}(\thetav) \prod_{i=1}^N q_{\phiv}(\hv_i | \xv_i)$. Let
$G_{\phiv}(\xv,\hv,\thetav) = \log p(\hv | \thetav) + \log p(\xv | \hv, \thetav)
- \log q_{\phiv}(\hv |\xv)$. The ELBO in Eq.~(\ref{eq:varBayes}) can be written
as
\begin{eqnarray}
\mathcal{L}(\phiv; \data) = \ep_q \left[ \log p_0(\thetav) + \! \sum_i \! G_{\phiv}(\xv_i, \hv_i, \thetav) - \log q_{\phiv}(\thetav) \right]. \nonumber
\end{eqnarray}
By using the equality $\nabla_{\phiv} q_{\phiv}(\Thetav) = q_{\phiv}(\Thetav) \nabla_{\phiv} \log q_{\phiv}(\Thetav)$, it can be shown that the gradient is
\begin{eqnarray}
\nabla_{\phiv} \mathcal{L}  = \ep_q \left[( \log p(\Thetav, \data) - \log q_{\phiv}(\Thetav) )  \nabla_{\phiv} \log q_{\phiv}(\Thetav) \right]. \nonumber
\end{eqnarray}
A naive Monte Carlo estimate of the gradient is
\begin{eqnarray}
\nabla_{\phiv} \mathcal{L}  \approx \frac{1}{L} \sum_{l=1}^L \left[( \log p(\Thetav^l, \data) - \log q_{\phiv}(\Thetav^l) )  \nabla_{\phiv} \log q_{\phiv}(\Thetav^l) \right], \nonumber
\end{eqnarray}
where $\Thetav^l \sim q_{\phiv}(\Thetav)$. Note that the sampling and the
gradient $ \nabla_{\phiv} \log q_{\phiv}(\Thetav^l)$ only depend on the
variational distribution, not the underlying model. However, the variance of
such an estimate can be too large to be useful. In practice, effective variance
reduction techniques are needed~\cite{Paisley:2012,Ranganath:2013}.

For continuous $\hv$, a reparameterization of the samples $\hv \sim
q_{\phiv}(\hv | \xv)$ can be derived using a differentiable transformation
$g_{\phiv}(\epsilonv, \xv)$ of a noise variable $\epsilonv$:
\begin{eqnarray}
\hv = g_{\phiv}(\epsilonv, \xv),~\textrm{where}~\epsilonv \sim p(\epsilonv).
\end{eqnarray}
This is known as {\it non-centered parameterization} (NCP) in
statistics~\cite{Papaspiliopoulos:2007}, while the original representation is
known as {\it centered parameterization} (CP).
A similar NCP reparameterization exists for the continuous $\thetav$:
\begin{eqnarray}
\thetav = f_{\phiv}(\zetav),~\textrm{where}~\zetav \sim p(\zetav).
\end{eqnarray}
Given a minibatch of data points $B_t$, we define $F_{\phiv}(\{\xv_i, \hv_i\}_{i\in B_t}, \thetav) = \frac{N}{|B_t|} \sum_{i \in B_t}  G_{\phiv}(\xv_i, \hv_i, \thetav ) + \log p_0(\thetav) - \log q_{\phiv}(\thetav)$.
Then, the Monte Carlo estimate of the variational lower bound is
\begin{eqnarray}
\mathcal{L}(\phiv; \data) \approx \frac{1}{L} \sum_{l=1}^L F_{\phiv}\left( \{\xv_i, g_{\phiv}(\epsilonv^l, \xv_i)\}_{i \in B_t}, f_{\phiv}(\zetav^l) \right),
\end{eqnarray}
where $\epsilonv^l \sim p(\epsilonv)$ and $\zetav^l \sim p(\zetav)$. This stochastic estimate can be maximized via gradient ascent methods.

It has been analyzed that CP and NCP possess complimentary
strengths~\cite{Papaspiliopoulos:2007}, in the sense that NCP is likely to work
when CP does not and conversely. An accompany paper~\cite{Kingma:2014b} to AEVB
analyzes the conditions for gradient-based samplers (e.g., HMC) whether NCP can be
effective or ineffective in reducing posterior dependencies; and it suggests to use
the interleaving strategy between centered and non-centered parameterization as
previously studied in~\cite{YuMeng:2011}. AEVB
has been extended to learn deep generative models~\cite{Rezende:2014} using the
similar reparameterization trick on continuous latent variables. However, AEVB cannot
be directly applied to deal with discrete variables.
In contrast, the work~\cite{Mnih:2014} presents a sophisticated method to reduce the variance of
the naive Monte Carlo estimate for deep autoregressive models; thus
it is applicable to both continuous and discrete latent variables.


\subsubsection{Stochastic Monte Carlo Methods}

The existing stochastic Monte Carlo methods can be generally grouped into three
categories, namely, stochastic gradient-based methods, the methods using
approximate MH test with randomly sampled mini-batches, and data augmentation.

{\bf Stochastic Gradient}: The idea of using gradient information to improve the
mixing rates has been systematically studied in various MC methods, including
Langevin dynamics and Hamiltanian dynamics~\cite{Neal:10}. For example, the
Langevin dynamics is an MCMC method that produces samples from the posterior by
means of gradient updates plus Gaussian noise, resulting in a proposal
distribution $p(\thetav_{t+1} | \thetav_t)$ by the following equation:
\begin{eqnarray}
\thetav_{t+1} = \thetav_t + \frac{\epsilon_t}{2} \left( \nabla_{\thetav} \log p_0(\thetav) + \nabla_{\thetav} \log p(\data | \thetav) \right) + \zeta_t,
\end{eqnarray}
where $\zeta_t \sim \mathcal{N}(0, \epsilon_t I)$ is an isotropic Gaussian noise
and $\log p(\data | \thetav) = \sum_i  \log p(\xv_i | \thetav)$ is the
log-likelihood of the full data set. The mean of the proposal distribution is in
the direction of increasing log posterior due to the gradient, while the
Gaussian noise will prevent the samples from collapsing to a single maximum. A
Metropolis-Hastings correction step is required to correct for discretisation
error~\cite{Roberts:02}.

The stochastic ideas have been successfully explored in these methods to develop
efficient stochastic Monte Carlo methods, including stochastic gradient Langevin
dynamics (SGLD)~\cite{Welling:icml11} and stochastic gradient Hamiltonian
dynamics (SGHD)~\cite{Chen:icml14}. For example, SGLD replaces the calculation
of the gradient over the full data set, with a stochastic approximation based on
a subset of data. Let $B_t$ be the subset of data points uniformly sampled from
the full data set at iteration $t$. Then, the gradient is approximated as:
\begin{eqnarray}\label{eq:stochastic-mc}
\nabla_{\thetav} \log p(\data | \thetav) \approx \frac{N}{|B_t|} \sum_{i \in B_t} \nabla_{\thetav} \log p(\xv_i | \thetav).
\end{eqnarray}
Note that SGLD doesn't use a MH correction step, as calculating the acceptance
probability requires use of the full data set. Convergence to the posterior is
still guaranteed if the step sizes are annealed to zero at a certain rate, as
rigorously justified in~\cite{Pillai:2014,Teh:2014sgld}.

To further improve the mixing rates, the stochastic gradient Fisher scoring
method~\cite{Ahn:2012} was developed, which represents an extension of the
Fisher scoring method based on stochastic gradients~\cite{Schraudolph:2007} by
incorporating randomness in a subsampling process. Similarly, exploring the
Riemannian manifold structure leads to the development of stochastic gradient
Riemannian Langevin dynamics (SGRLD)~\cite{Teh:nips13}, which performs SGLD on
the probability simplex space.


{\bf Approximate MH Test}: Another category of stochastic Monte Carlo methods
rely on approximate MH test using randomly sampled subset of data points, since
an exact calculation of the MH test in Eq.~(\ref{eq:MH}) scales linearly to the
data size, which is prohibitive for large-scale data sets. For example, the
work~\cite{Korattikara:2014} presents an approximate MH rule via sequential
hypothesis testing, which allows us to accept or reject samples with high
confidence using only a fraction of the data required for the exact MH rule. The
systematic bias and its tradeoff with variance were theoretically analyzed.
Specifically, it is based on the observation that the MH test rule in
Eq.~(\ref{eq:MH}) can be equivalently written as follows:
\begin{enumerate}
\item Draw $\gamma \sim \textrm{Uniform}[0,1]$ and compute:
\begin{eqnarray}
\mu_0 &=& \frac{1}{N} \log\left[ \gamma \frac{p_0(\Thetav_t) q(\Thetav^\prime | \Thetav_t) }{p_0(\Thetav^\prime) q(\Thetav_t | \Thetav^\prime)} \right] \nonumber \\
\mu &=& \frac{1}{N} \sum_{i=1}^N \ell_i,~ \textrm{where}~ \ell_i = \log \frac{p(\xv_i | \Thetav^\prime)}{p(\xv_i | \Thetav_t)}; \nonumber
\end{eqnarray}
\item If $\mu > \mu_0$ set $\Thetav_{t+1} \gets \Thetav^\prime$; otherwise $\Thetav_{t+1} \gets \Thetav_t$.
\end{enumerate}
Note that $\mu_0$ is independent of the data set, thus can be easily calculated.
This reformulation of the MH test makes it very easy to frame it as a
statistical hypothesis test, that is, given $\mu_0$ and a set of samples
$\{\ell_{t_1}, \dots, \ell_{t_n}\}$ drawn without replacement from the
population $\{\ell_1, \dots, \ell_N\}$, can we decide whether the population
mean $\mu$ is greater than or less than the threshold $\mu_0$? Such a test can
be done by increasing the cardinality of the subset until a prescribed
confidence level is reached. The MH test with approximate confidence intervals
can be combined with the above stochastic gradient methods (e.g., SGLD) to
correct their bias. The similar sequential testing ideas can be applied to Gibbs
sampling, as discussed in~\cite{Korattikara:2014}.

Under the similar setting of approximate MH test with subsets of data, the
work~\cite{Bardenet:2014} derives a new stopping rule based on some
concentration bounds (e.g., the empirical Bernstein bound~\cite{Bardenet:2013}),
which leads to an adaptive sampling strategy with theoretical guarantees on the
total variational norm between the approximate MH kernel and the target
distribution of MH applied to the full data set.

{\bf Data Augmentation}: 
The work~\cite{Maclaurin:2014} presents a Firefly Monte Carlo (FlyMC) method,
which is guaranteed to converge to the true target posterior. FlyMC relies on a
novel data augmentation formulation~\cite{DykMeng2001}. Specifically, let $z_i$
be a binary variable, indicating whether data $i$ is active or not, and
$B_i(\Thetav)$ be a strictly positive lower bound of the $i$th likelihood: $0 <
B_i(\Thetav) < L_i(\Thetav) \triangleq p(\xv_i | \Thetav)$. Then, the target
posterior $p(\Thetav | \data)$ is the marginal of the complete posterior with
the augmented variables $\Zv = \{z_i\}_{i=1}^N$:
\begin{eqnarray}
p(\Thetav, \Zv | \data) \propto p_0(\Thetav) \prod_{i=1}^N p(\xv_i | \Thetav) p(z_i | \xv_i, \Thetav),
\end{eqnarray}
where $p(z_i | \xv_i, \Thetav) = (1 - \gamma_i)^{z_i} \gamma_i^{(1-z_i)}$ and
$\gamma_i = B_i(\Thetav) / L_i(\Thetav)$. Then, we can construct a Markov chain
for the complete posterior by alternating between updates of $\Thetav$
conditioned on $\Zv$, which can be done with any conventional MCMC algorithm,
and updates of $\Zv$ conditioned on $\Thetav$, which can also been efficiently
done as 
we only need to re-calculate the likelihoods of the data points with active $z$
variables, thus effectively using a random subset of data points in each
iteration of the MC methods.

\subsection{Streaming Algorithms}

We can see that both (\ref{eq:stochastic-var}) and (\ref{eq:stochastic-mc}) need
to know the data size $N$, which renders them unsuitable for learning with
streaming data, where data comes in small batches without an explicit bound on
the total number as times goes along, e.g., tracking an aircraft using radar
measurements. This conflicts with the sequential nature of the Bayesian updating
procedure.
Specifically, let $B_t$ be the small batch at time $t$. Given the posterior at
time $t$, $p_t(\Thetav) := p(\Thetav | B_1, \dots, B_t)$, the posterior
distribution at time $t+1$ is
\begin{eqnarray}
p_{t+1}(\Thetav) := p(\Thetav | B_1, \dots, B_{t+1}) = \frac{p_t(\Thetav) p(B_{t+1} | \Thetav) }{ p(B_1, \dots, B_{t+1}) }.
\end{eqnarray}
In other words, the posterior at time $t$ is actually playing the role of a
prior for the data at time $t+1$ for the Bayesian updating. Under the
variational formulation of Bayes' rule, streaming RegBayes~\cite{Shi:2014} can
naturally be defined as solving:
\begin{eqnarray}
\min_{q(\Thetav) \in \prob} \KL(q(\Thetav) \Vert p_t(\Thetav)) + c \cdot \Omega(q(\Thetav); B_{t+1}),
\end{eqnarray}
whose streaming update rule can be derived via convex analysis under a quite general setting.

The sequential updating procedure is perfectly suitable for online learning with
data streams, where a revisit to each data point is not allowed. However, one
challenge remains on evaluating the posteriors. If the prior is conjugate to the
likelihood model (e.g., a linear Gaussian state-space model) or the state space
is discrete (e.g., hidden Markov models~\cite{Rabiner:89,Scott:2002}), then the
sequential updating rule can be done analytically, for example, Kalman
filters~\cite{Kalman:1960}. In contrast, many complex Bayesian models (e.g., the
models involving non-Gaussianity, non-linearity and high-dimensionality) do not
have closed-form expression of the posteriors. Therefore, it is computationally
intractable to do the sequential update.

\subsubsection{Streaming Variational Methods}
Various effects have been made to develop streaming variational Bayesian (SVB)
methods~\cite{Broderick:nips13}. Specifically, let $\mathcal{A}$ be a
variational algorithm that calculates the approximate posterior $q$: $q(\Thetav)
= \mathcal{A}( p(\Theta); B )$. Then, setting $q_0(\Thetav) = p_0(\Thetav)$, one
way to recursively compute an approximation to the posterior is
\begin{eqnarray}
p(\Thetav | B_1, \dots, B_{t+1}) \approx q_{t+1}(\Thetav) = \mathcal{A}( q_{t}(\Thetav), B_{t+1} ).
\end{eqnarray}
Under the exponential family assumption of $q$, the streaming update rule has some analytical form. 

The streaming RegBayes~\cite{Shi:2014} provides a Bayesian generalization of
online passive-aggressive (PA) learning~\cite{Crammer:2006}, when the posterior
regularization term is defined via the max-margin principle. The resulting
online Bayesian passive-aggressive (BayesPA) learning adopts a similar streaming
variational update to learn max-margin classifiers (e.g., SVMs) in the presence
of latent structures (e.g., latent topic representations). Compared to the
ordinary PA, BayesPA
is more flexible on modeling complex data. For example, BayesPA can discover
latent structures via a hierarchical Bayesian treatment as well as allowing for
nonparametric Bayesian inference to resolve the complexity of latent components
(e.g., using a HDP topic model to resolve the unknown number of topics).

\subsubsection{Streaming Monte Carlo Methods}

Sequential Monte Carlo (SMC)
methods~\cite{Andrieu:2010,Liu:1998,Arulampalam:2002} provide simulation-based
methods to approximate the posteriors for online Bayesian inference. SMC methods
rely on resampling and propagating samples over time with a large number of
particles. A standard SMC method would require the full data to be stored for
expensive particle rejuvenation to protect particles against degeneracy, leading
to an increased storage and processing bottleneck as more data are accrued. For
simple conjugate models, such as linear Gaussian state-space models, efficient
updating equations can be derived using methods like Kalman filters. For a
broader class of models, assumed density filtering
(ADF)~\cite{Lauritzen:1992,Opper:1999} was developed to extend the computational
tractability. Basically, ADF approximates the posterior distribution with a
simple conjugate family, leading to approximate online posterior tracking.
Recent improvements on SMC methods include the conditional density filtering
(C-DF) method~\cite{Guhaniyogi:2014}, which extends Gibbs sampling to streaming
data. C-DF sequentially draws samples from an approximate posterior distribution
conditioned on surrogate conditional sufficient statistics, which are
approximations to the conditional sufficient statistics using sequential samples
or point estimates for parameters along with the data. C-DF requires only data
at the current time and produces a provably good approximation to the target
posterior.

\subsection{Distributed Algorithms}


Recent progress has been made on both distributed variational and distributed Monte Carlo methods.

\subsubsection{Distributed Variational Methods}

If the variational distribution is in some parametric family (e.g., the
exponential family), the variational problem can be solved with generic
optimization methods. Therefore, the broad literature on distributed
optimization~\cite{Boyd:2011} provides rich tools for distributed variational
inference. However, the disadvantage of a generic solver is that it may fail to
explore the structure of Bayesian inference.

First, many Bayesian models have a nature hierarchy, which encodes rich
conditional independence structures that can be explored for efficient
algorithms, e.g., the distributed variational algorithm for
LDA~\cite{Zhai:2012}. Second, the inference procedure with Bayes' rule is
intrinsically parallelizable. Suppose the data $\data$ is split into
non-overlapping batches (often called shards), $B_1$, $\dots$, $B_M$. Then, the
Bayes posterior $p(\Thetav | \data) = \frac{p_0(\Thetav) \prod_{i=1}^M p(B_i |
\Thetav)}{p(\data)} $ can be expressed as
\setlength\arraycolsep{-5pt} \begin{eqnarray}
&& p(\Thetav | \data) 
= \frac{1}{C} \! \prod_{i=1}^M \frac{p_0(\Thetav)^{\frac{1}{M}} p(B_i | \Thetav)}{ p(B_i) } =  \frac{1}{C} \! \prod_{i=1}^M p(\Thetav | B_i),  \label{eqn:distributed-vb}
\end{eqnarray}
where $C = \frac{p(\data) }{\prod_{i=1}^M p(B_i)}$. Now, the question is how to
calculate the local posteriors (or subset posteriors) $p(\Thetav | B_i)$ as well
as the normalization factor.
The work~\cite{Broderick:nips13} explores this idea and presents a distributed
variational Bayesian method, which approximates the local posterior with an
algorithm~$\mathcal{A}$, that is, $p(\Thetav | B_i) \approx \mathcal{A}\left(
p_0(\Thetav)^{1/M}, B_i \right).$
Under the exponential family assumption of the prior and the approximate local
posteriors, the global posterior can be (approximately) calculated via density
product. However, the parametric assumptions may not be reasonable, and the
mean-field assumptions can get the marginal distributions right but not the
joint distribution.





\subsubsection{Distributed Monte Carlo Methods}\label{sec:distributed-monte-carlo}

For MC methods, if independent samples can be directly drawn from the posterior
or some proposals (e.g., using importance sampling), it will be
straightforward to parallelize, e.g., by running multiple independent samplers
on separate machines and then aggregating the samples~\cite{Wu:2012}. We
consider the more challenging cases, where directly sampling from the posterior
is intractable and MCMC methods are among the natural choices. There are two
groups of methods. One is to run multiple MCMC chains in parallel, and the other
is to parallelize a single MCMC chain. The ``multiple-chain" parallelism is
relatively straightforward if each single chain can be efficiently carried out
and an appropriate combination strategy is adopted~\cite{Gelman:1992,Wu:2012}.
However, in Big data applications a single Markov chain itself is often
prohibitively slow to converge, due to the massive data sizes or extremely
high-dimensional sample spaces.
Below, we focus on the methods that parallelize a single Markov chain, under 
three major categories.


{\bf Blocking}: Methods in this category 
let each computing unit (e.g., a CPU processor or a GPU core) to perform a part
of the computation at each iteration. For example, they independently evaluate
the likelihood for each shard across multiple units and combine the local
likelihoods with the prior on a master unit to get estimates of the global
posterior~\cite{Suchard:2010}. Another example is that each computing unit is
responsible for updating a part of the state space~\cite{Wilkinson:2004}.
These methods involve extensive communications and being problem specific.

In these methods several computing units collaborate to obtain a draw from the
posterior.
In order to effectively split the likelihood evaluation or the state space
update over multiple computing units, it is important to explore the conditional
independence (CI) structure of the model. Many hierarchical Bayesian models
naturally have the CI structure (e.g., topic models), while some other models
need some transformation to introduce CI structures that are appropriate for
parallelization~\cite{Williamson:2013}.

{\bf Divide-and-Conquer}: Methods in this category avoid extensive
communication among machines by running independent MCMC chains on each
shard and aggregating samples drawn from local posteriors via a single
communication. Aggregating the local samples is the key step,
with a lot of recent progress. For example, the consensus
Monte Carlo~\cite{Scott:2013}~directly combines local samples by a weighted
average, which is valid under an implicit Gaussian assumption while lacking of
guarantees for non-Gaussian cases;~\cite{Neiswanger:2014} approximates each
local posterior with either an explicit Gaussian or a Gaussian-kernel KDE so
that combination follows an explicit density product; \cite{Wang:2013} builds
upon the KDE idea one step further by representing the discrete KDE as a
continuous Weierstrass transform; and
~\cite{Minsker:2014} proposes to calculate the geometric median of local
posteriors (or M-posterior), which is provably robust to the presence of
outliers. The M-posterior is approximately solved by the Weiszfeld's
algorithm~\cite{Beck:2014} by embedding the local posteriors in a reproducing
kernel Hilbert space.

The potential drawback of these embarrassingly parallel MCMC sampling is that if
the local posteriors differ significantly, perhaps due to noise or non-random
partitioning of the dataset across nodes, the final combination stage can result
in inaccurate global posterior. The recent work~\cite{Minjie:2014} presents a
context aware distributed Bayesian posterior sampling method to improve
inference quality. By allowing nodes to effectively and efficiently share
information with each other, each node will eventually draw samples from a more
accurate approximate full posterior, and therefore no long needs any
combination.

\begin{figure}[t]\vspace{-.2cm}
\centering
\includegraphics[width = .32\textwidth]{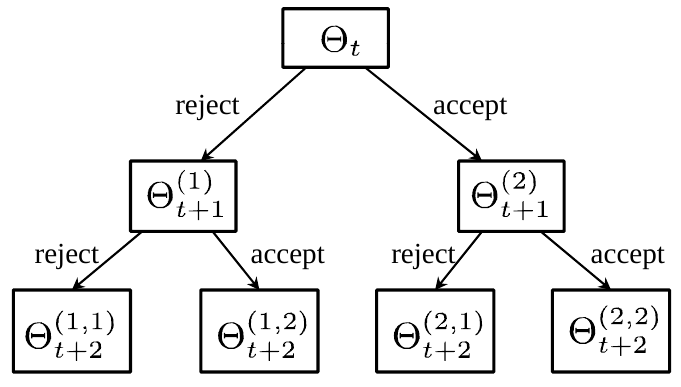}\vspace{-.2cm}
\caption{The possible outcomes in two iterations of a Metropolis-Hastings sampler.}\label{fig:prefetching}\vspace{-.4cm}
\end{figure}

{\bf Prefetching}: The idea of prefetching is to make use of parallel processing
to calculate multiple likelihoods ahead of time, and only use the ones which are
needed. Consider a generic random-walk metropolis-Hastings algorithm at time
$t$. The subsequent steps can be represented by a binary tree, where at each
iteration a single new proposal is drawn from a proposal distribution and
stochastically accepted or rejected. So, at time $t + n$ the chain has $2^n$
possible future states, as illustrated in Fig.~\ref{fig:prefetching}. The
vanilla version of prefetching speculatively evaluates all paths in this binary
tree~\cite{Brockwell:2006}. Since only one path of these will be taken, with $M$
cores, this approach achieves a speedup of $\log_2 M$ with respect to single
core execution, ignoring communication overheads. More efficient prefetching
approaches have been proposed in~\cite{Angelino:2014} and \cite{Strid:2010} by
better guessing the probabilities of exploration of both the acceptance and the
rejection branches at each node. The recent work~\cite{Banterle:2014} presents a
delayed acceptance strategy for MH testing, which can be used to improve the
efficiency of prefetching.

As a special type of MCMC, Gibbs sampling methods naturally follow a blocking
scheme by iterating over some partition of the variables. The early asynchronous
Gibbs sampler~\cite{Geman:1984} 
is highly parallel by sampling all variables
simultaneously on separate processors. However, the extreme parallelism comes at
a cost, e.g., the sampler may not converge to the correct stationary
distribution in some cases~\cite{Gonzalez:2011}. The work~\cite{Gonzalez:2011}
develops various variable partitioning strategies to achieve fast
parallelization, while maintaining the convergence to the target posterior, and
the work~\cite{Johnson:2013} analyzes the convergence and correctness of the
asynchronous Gibbs sampler (a.k.a, the Hogwild parallel Gibbs sampler) for
sampling from Gaussian distributions. Many other parallel Gibbs sampling
algorithms have been developed for specific models. For example, various
distributed Gibbs samplers~\cite{Newman:2007,Smola:vldb10,Ahmed:2012,Liu:2011,Chen:2016}
have been developed for the vanilla LDA, \cite{Chen:2013} develops a distributed
Gibbs sampler via data augmentation to learn large-scale topic graphs with a
logistic-normal topic model, and parallel algorithms for DP mixtures have been
developed by introducing auxiliary variables for additional CI
structures~\cite{Williamson:2013}, while with the potential risk of causing
extremely imbalanced partitions~\cite{Gal:parallDP}.

Note that the stochastic methods and distributed computing are not exclusive.
Combing both often leads to more efficient solutions. For example, for
optimization methods, parallel SGD methods have been extensively
studied~\cite{Zinkevich:2010,Niu:2011}. In particular, \cite{Niu:2011} presents
a parallel SGD algorithm without locks, called Hogwild!, where multiple
processors are allowed equal access to the shared memory and are able to update
individual components of memory at will. Such a scheme is particularly suitable
for sparse learning problems.
For Bayesian methods, the distributed stochastic gradient Langevin dynamics
(D-SGLD) method has been developed in~\cite{Ahn:2014} and further improved for topic models in~\cite{Yang:KDD16}.

\section{Tools, Software and Systems}

Though stochastic algorithms are easy to implement, distributed methods often
need a careful design of the system architectures and programming libraries. For
system architectures, we may have a shared memory computer with many cores, a
cluster with many machines interconnected by network (either commodity or
high-speed), or accelerating hardware like graphics processing units (GPUs) and
field-programmable gate arrays (FPGAs).
We now review
the distributed programming frameworks suitable for various system architectures
and existing tools for Bayesian inference.

\subsection{System Primitives}

Every architecture has its low-level libraries, in which the parallel computing units
(e.g., threads, machines, or GPU cores) are explicitly visible to the programmer.

\textbf{Shared Memory Computer}: A shared memory computer
passes data from one CPU core to another by simply storing it into the main
memory. Therefore, the communication latency is low. It is also easy to program
and acquire. Meanwhile it is prevalent---it is the basic component of large
distributed clusters and host of GPUs or other accelerating hardware.
Due to these reasons,
writing a multi-thread program is usually the first step towards large-scale
learning. However, its drawbacks include limited memory/IO capacity and
bandwidth, and restricted scalability, which can be addressed by distributed
clusters.

Programmers work with threads in a shared memory setting. A threading library
supports: 1) spawning a thread and wait it to complete;
2) synchronization: method to prevent conflict access of resources, such as
locks; 3) atomic: operations, such as increment that can be executed in parallel
safely.
Besides threads and locks, there are alternative programming frameworks. 
For example, Scala uses \emph{actor}, which responds to a message that it
receives; Go uses \emph{channel}, which is a multi-provider, multi-consumer
queue. There are also libraries automating specific parallel pattern, e.g.,
OpenMP~\cite{OpenMP} supports parallel patterns like parralel for or reduction,
and synchronization patterns like barrier; TBB~\cite{TBB} has pipeline,
lightweight green threads and concurrent data structures. Choosing right programming models sometimes can
simplify the implementation.

\textbf{Accelerating Hardware}:
GPUs are self-contained parallel computational devices that can be housed in
desktop or laptop computers. A single GPU can provide floating
operations per second (FLOPS) performance as good as a small cluster. Yet
compared to conventional multi-core processors, GPUs are cheap, easily
accessible, easy to maintain, easy to code, and dedicated local devices with low
power consumption. GPUs follow a single instruction multiple data (SIMD)
pattern, i.e., a single program will be executed on all cores given different
data. This pattern is suitable for many ML applications.
However, GPUs may be limited due to: 1) small memory capacity; 2) restricted
SIMD programming model; and 3) high CPU-GPU or GPU-GPU communication latency.

Many Bayesian inference methods have been accelerated with GPUs. For
example,~\cite{Suchard:2010} adopts GPUs to parallelize the likelihood
evaluation in MCMC;~\cite{Lee:2010} provides GPU parallelization for
population-based MCMC methods~\cite{Jasra:2007} as well as SMC
samplers~\cite{Moral:2006}; and~\cite{Beam:2014} uses GPU computing to develop
fast Hamiltonian Monte Carlo methods. For variational Bayesian
methods,~\cite{Yan:2009} demonstrates an example of using GPUs to accelerate the
collapsed variational Bayesian algorithm for LDA. More recently, SaberLDA~\cite{Li:2016} implements a sparsity-aware sampling algorithm on GPU, which scales sub-linearly with the number of topics.
BIDMach~\cite{Canny:2013} is a distributed GPU framework for machine learning,
In particular, BIDMach LDA with a single GPU
is able to learn faster than the state-of-the-art CPU based LDA implementation~\cite{Ahmed:2012},
which use 100 CPUs.

Finally, acceleration with other hardware (e.g., FPGAs) has also been investigated~\cite{Chau:2013}.

\textbf{Distributed Cluster}:
For distributed clusters, a low-level framework should allow users to do: 1)
{\it Communication}: sending and receiving data from/to another machine or a
group of machines; 2) {\it Synchronization}: synchronize the processes; 3) {\it
Fault handling}: decide what to do if a process/machine breaks down. For
example, MPI provides a set of primitives including {\it send}, {\it receive},
{\it broadcast} and {\it reduce} for communication. MPI also provides
synchronization operations, such as {\it barrier}. MPI handles fault by simply
terminating all processes. MPI works on various network infrastructures, such as
ethernet or Infiniband. Besides MPI, there are other frameworks that support
communication, synchronization and fault handling, such as 1) message queues,
where processes can put and get messages from globally shared message queues; 2)
remote procedural calls (RPCs), where a process can invoke a procedure in
another process, passing its own data to that remote procedure, and finally get
execution results. MrBayes~\cite{Huelsenbeck:2003,Altekar:2004} provides a
MPI-based parallel algorithm for Metropolis-coupled MCMC for Bayesian
phylogenetic inference.

Programming with system primitive libraries are most flexible and lightweight.
However for sophisticated applications, which may require asynchronous
execution, need to modify the global parameters while running, or need many
parallel execution blocks, it would be painful and error prone to write the
parallel code using the low-level system primitives.
{Below, we review some high-level distributed computing frameworks, which automatically execute the user declared tasks on desired architectures.}
We refer the readers to \cite{Bekkerman:2011} for more details on GPUs, MapReduce, and some other examples (e.g., parallel online learning).

\begin{figure}[t]
\centering
\includegraphics[width = .48\textwidth]{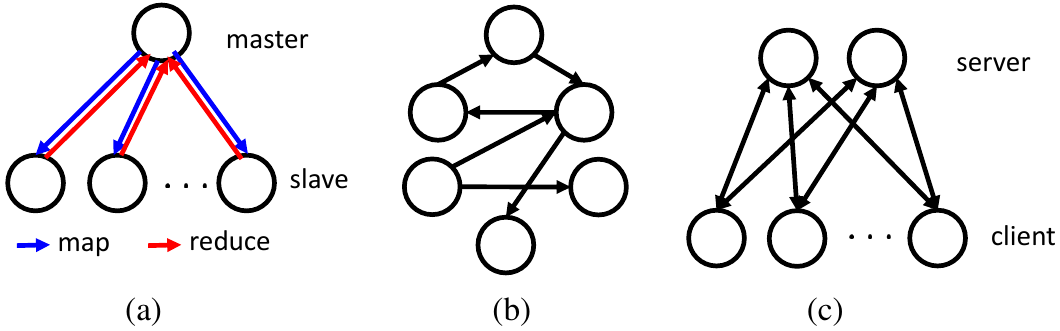}\vspace{-.3cm}
\caption{Various architectures: (a) MapReduce/Spark; (b) Pregel/GraphLab; (c) Parameter servers.} \label{fig:frameworks}\vspace{-.4cm}
\end{figure}

\vspace{-.1cm}
\subsection{MapReduce and Spark}
\vspace{-.05cm}

MapReduce~\cite{Dean:2004} is a distributed computing framework for key-value
stores. It reads key-value stores from disk, performs some transformations to
these key-value stores in parallel, and writes the final results to disk. A
typical MapReduce cycle involves the steps: (1) 
Spawn some workers on all machines; (2) Workers read input key-value pairs in parallel from a distributed file system;
(3) \textit{Map:} Pass each key-value pair to a user defined function, which will generate some intermediate key-value pairs;
(4) According to the key, hash the intermediate key-value pairs to all machines, then merge key-value pairs that have the same key, result with (key, list of values) pairs;
(5) \textit{Reduce:} In parallel, pass each (key, list of values) pairs to a user defined function, which will generate some output key-value pairs;
and (6) Write output key-value pairs to the file system.

There are two user defined functions, {\it mapper} and {\it reducer}. For ML, a
key-value store is often data samples, mapper is often used for computing latent
variables, likelihoods or gradients for each data sample, and reducer is often
used to aggregate the information from each data sample, where the information
can be used for estimating parameters or checking convergence.
\cite{Chu:2007} discusses a number of ML algorithms on MapReduce, including
linear regression, naive Bayes, neural networks, PCA, SVM, etc.
Mahout~\cite{Mahout} is a ML package built upon Hadoop, an open source
implementation of MapReduce. Mahout provides collaborative filtering,
classification, clustering, dimensionality reduction and topic modeling
algorithms.~\cite{Zhai:2012} is a MapReduce based LDA. However, a major drawback
of MapReduce is that it needs to read the data from disk at \emph{every
iteration}. The overhead of reading data becomes dominant for many iterative ML
algorithms as well as interactive data analysis tools~\cite{Zaharia:2010}.

Spark~\cite{Zaharia:2010} is another framework for distributed ML methods that
involve iterative jobs. The core of Spark is resilient data sets (RDDs), which
is essentially a dataset distributed across machines. RDD can be stored either
in memory or disk: Spark decides it automatically, and users can provide hints
to Spark which to store in memory. This avoids reading the dataset at every
iteration. Users can perform \emph{parallel operations} to RDDs, which will
transform a RDD to another. Available parallel operations are like {\it foreach}
and {\it reduce}. We can use foreach to do the computation for each data, and
use reduce to aggregate information from data. Because parallel operations are
just a parallel version of the corresponding serial operations, a Spark program
looks almost identical to its serial counterpart.
Spark can outperform Hadoop for iterative ML jobs by 10x, and is able to
interactively query a 39GB dataset in 1 second~\cite{Zaharia:2010}.

\vspace{-.1cm}
\subsection{Iterative Graph Computing}
\vspace{-.05cm}

Both MapReduce and Spark have a star architecture as in
Fig.~\ref{fig:frameworks} (a), where only master-slave communication is
permitted; they do not allow one key-value pair to interact with another, e.g.,
reading or modifying the value of another key-value pair. The interaction is
necessary for applications like PageRank, Gibbs sampling, and variational Bayes
optimized by coordinate descent, all of which require variables to get their own
values based on other related variables. Hence there comes graph computing,
where the computational task is defined by a sparse graph that specifies the
data dependency, as shown in Fig.~\ref{fig:frameworks} (b).

Pregel~\cite{Malewicz:2010} is a bulk synchronous parallel (BSP) graph computing
engine. The computation model is a sparse graph with data on vertices and edges,
where each vertex receives all messages sent to it in the last iteration;
updates data on the vertex based on the messages; and sends out messages along
adjacent edges.
For example, Gibbs sampling can be done easily by sending the vertex statistics
to adjacent vertices and then the conditional probability can be computed.
GPS~\cite{Salihoglu:gps2013} is an open source implementation of Pregel
with new features (e.g., dynamic graph repartition).

GraphLab~\cite{Low:2010} is a more sophisticated graph computing engine that
allows asynchronous execution and flexible scheduling. A GraphLab iteration
picks up a vertex $v$ in the task queue; and
passes the vertex to a user defined function, which may modify the data on the
vertex, its adjacent edges and vertices, and finally may add its adjacent
{vertices} to the task queue. Note that several nodes can be evaluated in
parallel as long as they do not violate the consistency guarantee {which ensures
that GraphLab is equivalent with some serial algorithm}. It has been used to
parallelize a number of ML tasks, including matrix factorization, Gibbs sampling
and Lasso~\cite{Low:2010}. \cite{Zhu:ACL13} presents a distributed Gibbs sampler
on GraphLab for an improved sLDA model using RegBayes. Several other graph
computing engines have been developed. For example, GraphX~\cite{Xin:2013} is an
extension of Spark for graph computing; and GraphChi~\cite{Kyrola:2012} is a
disk based version of GraphLab.


\vspace{-.1cm}
\subsection{Parameter Servers}
\vspace{-.05cm}

All the above frameworks restrict the communication between workers. For
example, MapReduce and Spark don't allow communication between workers, while
Pregel and GraphLab only allow vertices to communicate with adjacent nodes. On
the other side, many ML methods follow a pattern that: (1)
Data are partitioned on many workers; (2) There are some shared global parameters (e.g., the model weights in a
gradient descent method or the topic-word count matrix in the collapsed Gibbs
sampler for LDA~\cite{Griffiths:04}); and (3) Workers fetch data and update (parts of) global parameters based on
their local data (e.g., using the local gradients or local sufficient
statistics).
Though it is straightforward to implement on shared memory computers, 
it is rather difficult in a distributed setting. The goal of parameter servers
is to provide a distributed data structure for parameters.

A parameter server is a key-value store (like a hash map), accessible for all
workers. It supports for \texttt{get} and \texttt{set} (or \texttt{update}) for
each entry. In a distributed setting, both server and client consist of many
nodes (see Fig.~\ref{fig:frameworks} (c)). Memcached~\cite{Memcached} is an in
memory key-value store that provides \texttt{get} and \texttt{set} for arbitrary
data. However it doesn't have a mechanism to resolve conflicts raised by
concurrent access, e.g. concurrent writes for a single entry. Applications
like~\cite{Smola:vldb10} require to lock the global entry while updating, which
leads to suboptimal performance. Piccolo~\cite{Power:2010} addresses this by
introducing user-defined accumulations, which correctly address concurrent
\texttt{update}s to the same key. Piccolo has a set of built-in user defined
accumulations such as summation, multiplication, and min/max.


One important tradeoff made by parameter servers is that they sacrifice
consistency for less latency---\texttt{get} may not return the most recent
value, so that it can return immediately without waiting for most recent updates
to reach the server. While this improves the performance significantly, it can
potentially slow down convergence due to outdated parameters. \cite{Ho:2013}
proposed Stale Synchronous Parallel (SSP), where the staleness of parameters is
\emph{bounded} and the fastest worker can be ahead of the slowest one by no more
than $\tau$ iterations, where $\tau$ can be tuned to get a fast convergence as
well as low waiting time. 
Petuum~\cite{Dai:2013} is a SSP based parameter server. \cite{Li:paramserv2014}
proposed communication-reducing improvements, including key caching, message
compression and message filtering, and it also supports elastically adding and
removing both server and worker nodes.

Parameter servers have been deployed in learning very large-scale logistic
regression~\cite{Li:paramserv2014}, deep networks~\cite{Dean:2012},
LDA~\cite{Ahmed:2012, Li:2014} and Lasso~\cite{Dai:2013}.
\cite{Li:paramserv2014} learns a 2000-topic LDA with 5 billion documents and 5
million unique tokens on 6000 machines in 20 hours. Yahoo! LDA~\cite{Ahmed:2012}
has a parameter server designed specifically for Bayesian latent variable models
and it is the fastest available LDA software.
There are a bunch of distributed topic modeling softwares based on Yahoo! LDA,
including \cite{Zhu:KDD13} for MedLDA and~\cite{Chen:2013} for correlated topic
models.

\vspace{-.1cm}
\subsection{Model Parallel Inference}
\vspace{-.05cm}

MapReduce, Spark and Parameter servers take the \emph{data-parallelism}
approach, where \emph{data} are partitioned across machines and computations are
performed on each node given a copy of the globally shared \emph{model}.
However, as the model size rapidly grows (i.e., the large $M$ challenge),
the models cannot fit in a single computer's memory.
Model-parallelism addresses this challenge by partitioning the \emph{model} and
storing a part of the model on each node.
Then, partial updates (i.e., the updates of model parts) are carried out on each
node. Benefits of model-parallelism include large model sizes, flexibility to
focus workers on fastest-converging parameters, and more accurate convergence
because no delayed update is involved.

STRADS~\cite{Lee:2014} provides primitives for model-parallelism and it handles
the distributed storage of model and data automatically. STRADS requires that a
partial update
could be computed using just the model part together with data. Users writes
\textit{schedule} that assigns model sets to workers, \textit{push} that
computes the partial updates for model, \textit{pop} that applies updates to
model. An automatic \textit{sync} primitive will ensure that users always get
the latest model.
As a concrete example, \cite{Zheng:arxiv2014} demonstrates a model parallel LDA,
in which both data and model are partitioned by vocabulary. In each iteration, a
worker only samples latent variables and updates the model related to the
vocabulary part assigned to it. The model then rotates between workers, until a
full cycle is completed. Unlike data parallel
LDA~\cite{Smola:vldb10,Ahmed:2012,Newman:2007}, the sampler always uses the
latest models and no read-write lock is needed on models, thereby leading to
faster convergence than data-parallel LDAs.

Note that model-parallelism is not a replacement but a complement of
data-parallelism. For example, \cite{Wang:2014towards} showed a two layer LDA
system, where layer 1 is model-parallelism and layer 2 consists of several local
model-parallelism clusters performing asynchronous updates on an globally
distributed model.

\vspace{-.1cm}
\section{Conclusions and Perspectives}
\vspace{-.05cm}

We present a survey of recent advances on big learning with Bayesian methods,
including Bayesian nonparametrics, regularized Bayesian inference, and scalable
inference algorithms and the systems based-on stochastic subsampling or distributed computing.
It is helpful to note that our review is not exhaustive. In fact, big learning has
attracted intense interest with active research spanning diverse fields,
including machine learning, databases, parallel and distributed systems, and
programming languages.





As reviewed above, big learning with Bayesian methods has achieved substantial progress. However, considerable challenges still remain. We briefly discuss several directions that are of promise for future investigation.
First, Bayesian methods have the advantage to incorporate prior knowledge for efficient learning,
especially for the scenarios where a large number of training data is lacking, and characterize uncertainty. For instance,
the recent work~\cite{Lake:BPL15} demonstrates an example for the challenging task of one-shot learning,
which achieves human-level performance by encoding the
domain knowledge as a hierarchical Bayesian model.
In contrast, deep learning methods~\cite{LeCun-nature2015} stand at the other end of the spectrum---they are often learned
in an end-to-end manner by feeding a large set of training data, and they often do not represent the uncertainty
in the structure or parameters of the neural networks.
A natural and important question that remains under-addressed is how to conjoin the flexibility of deep learning
and the learning efficiency of Bayesian methods for robust learning.
Another related important question is how to effectively collect domain knowledge and incorporate it into
the modeling and inference process. The work~\cite{Mei:robustBayes14}
has demonstrated an example that selectively incorporates the noisy knowledge collected from
crowds for robust Bayesian inference, but much more are left unexplored.

Second, one of the lessons we learn from big learning is that the best predictive performance
is often obtained by building a highly flexible model (e.g., deep neural networks~\cite{LeCun-nature2015}).
Although nonparametric Bayesian techniques are powerful in theory to represent flexible models and
automatically infer their complexity from an unbounded space,
there is still a large gap in practice, with very few real applications. Most of the
evaluations are proof-of-concepts by being hindered on small-scale problems or those with relatively simple structures.
For example, although some attempts have demonstrated that a cascade IBP can be applied to infer the structure of a sparse deep belief network~\cite{Adams:2011},
these results are preliminary and can only learn toy network structures. It needs further study
on how to learn the structure of a sophisticated network with state-of-the-art performance.
In order to fill up the practical gap of nonparametric models, we need to
develop the algorithms that are accurate and scalable as well as
the theory of defining flexible nonparametric processes that
can properly consider the rich structures in various domains.

Third, a more powerful way of composing Bayesian models is offered by
probabilistic programming\footnote{http://probabilistic-programming.org},
which uses general-purpose computer programs to represent probabilistic models and
automates the inference procedure by building a universal engine. Several probabilistic programming languages
have been developed, including BUGS\footnote{http://www.mrc-bsu.cam.ac.uk/software/bugs/},
Stan\footnote{http://mc-stan.org/}, BLOG\footnote{https://bayesianlogic.github.io/},
Church\footnote{https://projects.csail.mit.edu/church/wiki/Church} and Infer.Net\footnote{http://research.microsoft.com/en-us/um/cambridge/projects/infernet/}.
However, scalable inference is still a considerable challenge for these languages.
The existing platforms for Bayesian inference do not well support the advanced deep models
and the recent scalable algorithms in distributed/stochastic settings. They do not well support the
accelerating hardware (e.g., GPUs and FPGAs) either.
In fact, the existence of user-friendly platforms (e.g., Tensorflow~\cite{tensorflow2015-whitepaper}, Theano~\cite{2016arXiv160502688short} and Caffe~\cite{jia2014caffe})
has significantly boosted the applications of deep learning in industry.
It will be very useful to fill up this gap for Bayesian methods, which can allow for rapid prototyping and testing of different models,
therefore motivating wider adoption of Bayesian methods. Edward\footnote{http://edwardlib.org/} is a recent system that builds on Tensorflow for scalable Bayesian inference, but much work needs to be done.

Finally, the current machine learning methods in general still require considerable human expertise in devising
appropriate features, priors, models, and algorithms. Much work has to be done in order to make
ML more widely used and eventually become a common part of our day to day tools in data sciences.
Along this line, several promising projects have been started. Google prediction API is one of
the earliest efforts that try to make ML accessible for beginners by
providing easy-to-use service. Microsoft AzureML takes a
similar approach by providing a visual interface to help design experiments.
SystemML~\cite{Ghoting:2011} provides an R-like declarative language to specify
ML tasks based on MapReduce, and MLBase~\cite{Kraska:2013} further improves it
by providing learning-specific optimizer that transforms a declarative task into
a sophisticated learning plan. Finally, Automated Statistician
(AutoStat)~\cite{Lloyd:2014} aims to automate the process of statistical
modeling, by using Bayesian model selection strategies to automatically choose
good models/features and to interpret the results in easy-to-understand
ways, in terms of automatically generated reports. Though still at a very early stage,
such efforts would have a tremendous impact on the fields that currently rely on expert
statisticians, ML researchers, and data scientists.



%
\vspace{-.2cm}
\section*{Acknowledgements}
\vspace{-.13cm}
{\small The work is supported by National 973 Projects
(2013CB329403), NSF of China Projects (61322308, 61332007), and Tsinghua Initiative Scientific Research Program
(20121088071).}

\vspace{-.2cm}
\bibliographystyle{abbrv}
\bibliography{bigBayes}\vspace{-1.6cm}

\begin{thebibliography}{100}

\bibitem{Mahout}
Apache mahout: https://mahout.apache.org/.

\bibitem{LSHTC}
http://lshtc.iit.demokritos.gr/.

\bibitem{Memcached}
http://memcached.org.

\bibitem{TBB}
https://www.threadingbuildingblocks.org/.

\bibitem{ImageNet}
http://www.image-net.org/about-overview.

\bibitem{OpenMP}
http://www.openmp.org.

\bibitem{tensorflow2015-whitepaper}
M.~Abadi, A.~Agarwal, P.~Barham, E.~Brevdo, Z.~Chen, C.~Citro, G.~S. Corrado,
  A.~Davis, J.~Dean, M.~Devin, S.~Ghemawat, I.~Goodfellow, A.~Harp, G.~Irving,
  M.~Isard, Y.~Jia, R.~Jozefowicz, L.~Kaiser, M.~Kudlur, J.~Levenberg,
  D.~Man\'{e}, R.~Monga, S.~Moore, D.~Murray, C.~Olah, M.~Schuster, J.~Shlens,
  B.~Steiner, I.~Sutskever, K.~Talwar, P.~Tucker, V.~Vanhoucke, V.~Vasudevan,
  F.~Vi\'{e}gas, O.~Vinyals, P.~Warden, M.~Wattenberg, M.~Wicke, Y.~Yu, and
  X.~Zheng.
\newblock {TensorFlow}: Large-scale machine learning on heterogeneous systems,
  2015.
\newblock Software available from tensorflow.org.

\bibitem{Adams:2011}
R.~Adams, H.~Wallach, and Z.~Ghahramani.
\newblock Learning the structure of deep sparse graphical models.
\newblock In {\em AISTATS}, 2010.

\bibitem{Ahmed:2012}
A.~Ahmed, M.~Aly, J.~Gonzalez, S.~Narayanamurthy, and A.~Smola.
\newblock Scalable inference in latent variable models.
\newblock In {\em WSDM}, 2012.

\bibitem{Ahn:2012}
S.~Ahn, A.~Korattikara, and M.~Welling.
\newblock Bayesian posterior sampling via stochastic gradient fisher scoring.
\newblock In {\em ICML}, 2012.

\bibitem{Ahn:2014}
S.~Ahn, B.~Shahbaba, and M.~Welling.
\newblock Distributed stochastic gradient {MCMC}.
\newblock In {\em ICML}, 2014.

\bibitem{Aitchison:80}
J.~Aitchison and S.~M. Shen.
\newblock Logistic-normal distributions: Some properties and uses.
\newblock {\em Biometrika}, 67(2):261--272, 1980.

\bibitem{Altekar:2004}
G.~Altekar, S.~Dwarkadas, J.~Huelsenbeck, and F.~Ronquist.
\newblock Parallel {Metropolis} coupled {Markov} chain {Monte Carlo for
  Bayesian} phylogenetic inference.
\newblock {\em Bioinformatics}, 20(3):407--415, 2004.

\bibitem{Amari:1998}
S.~Amari.
\newblock Natural gradient works efficiently in learning.
\newblock {\em Neural Comput.}, 10:251--276, 1998.

\bibitem{Andrieu:2010}
C.~Andrieu, A.~Doucet, and R.~Holenstein.
\newblock Particle {Markov} chain {Monte Carlo} methods.
\newblock {\em J. R. Stat. Soc., Ser B}, 72(3):269--342, 2010.

\bibitem{Andrieu:2003}
C.~Andrieu, N.~D. Freitas, A.~Doucet, and M.~I. Jordan.
\newblock An introduction to {MCMC} for machine learning.
\newblock {\em Machine Learning}, 50:5--43, 2003.

\bibitem{Angelino:2014}
E.~Angelino, E.~Kohler, A.~Waterland, M.~Seltzer, and R.~P. Adams.
\newblock Accelerating {MCMC} via parallel predictive prefetching.
\newblock {\em arXiv:1403.7265}, 2014.

\bibitem{Antoniak:74}
C.~Antoniak.
\newblock Mixture of {Dirichlet} process with applications to {Bayesian}
  nonparametric problems.
\newblock {\em Ann. Stats.}, (273):1152--1174, 1974.

\bibitem{Arulampalam:2002}
M.~Arulampalam, S.~Maskell, N.~Gordon, and T.~Clapp.
\newblock A tutorial on particle filters for online nonlinear/non-gaussian
  {Bayesian} tracking.
\newblock {\em IEEE Trans. Signal Process.}, 50(2):174--188, 2002.

\bibitem{Banterle:2014}
M.~Banterle, C.~Grazian, and C.~P. Robert.
\newblock Accelerating {Metropolis-Hastings} algorithms: Delayed acceptance
  with prefetching.
\newblock {\em arXiv:1406.2660}, 2014.

\bibitem{Bardenet:2014}
R.~Bardenet, A.~Doucet, and C.~Holmes.
\newblock Towards scaling up {Markov chain Monte Carlo}: an adaptive
  subsampling approach.
\newblock In {\em ICML}, 2014.

\bibitem{Bardenet:2013}
R.~Bardenet and O.-A. Maillard.
\newblock Concentration inequalities for sampling without replacement.
\newblock {\em arXiv:1309.4029}, 2013.

\bibitem{Beal:iHMM07}
M.~Beal, Z.~Ghahramani, and C.~Rasmussen.
\newblock The infinite hidden markov model.
\newblock In {\em NIPS}, 2002.

\bibitem{Beal:2003}
M.~J. Beal.
\newblock Variational algorithms for approximate {Bayesian} inference.
\newblock {\em PhD Thesis, University of Cambridge}, 2003.

\bibitem{Beam:2014}
A.~L. Beam, S.~K. Ghosh, and J.~Doyle.
\newblock Fast {Hamiltonian Monte Carlo} using {GPU} computing.
\newblock {\em arXiv:1402.4089}, 2014.

\bibitem{Beck:2014}
A.~Beck and S.~Sabach.
\newblock Weiszfeld¡¯s method: Old and new results.
\newblock {\em J. of Opt. Theory and Applications}, 2014.

\bibitem{Bekkerman:2011}
R.~Bekkerman, M.~Bilenko, and J.~Langford.
\newblock {\em Scaling up machine learning: Parallel and distributed
  approaches}.
\newblock Cambridge University Press, 2011.

\bibitem{Bengio:nips10}
S.~Bengio, J.~Weston, and D.~Grangier.
\newblock Label embedding trees for large multi-class tasks.
\newblock In {\em NIPS}, 2010.

\bibitem{Bengio:2012}
Y.~Bengio, A.~Courville, and P.~Vincent.
\newblock {Representation Learning: A Review and New Perspectives}.
\newblock {\em IEEE Trans. on PAMI}, 35(8):1798--1828, 2013.

\bibitem{Bialek:2001}
W.~Bialek, I.~Nemenman, and N.~Tishby.
\newblock Predictability, complexity and learning.
\newblock {\em Neural Comput.}, 13:2409--2463, 2001.

\bibitem{Bishop:PRML}
C.~M. Bishop.
\newblock {\em Pattern Recognition and Machine Learning}.
\newblock Springer, 2006.

\bibitem{Blei:2010}
D.~Blei and P.~Frazier.
\newblock Distance dependent {Chinese} restaurant processes.
\newblock In {\em ICML}, 2010.

\bibitem{Blei:hdp06}
D.~Blei and M.~Jordan.
\newblock Variational inference for dirichlet process mixtures.
\newblock {\em Bayesian Analysis}, 1:121--144, 2006.

\bibitem{Blei:06}
D.~Blei and J.~Lafferty.
\newblock Correlated topic models.
\newblock In {\em NIPS}, 2006.

\bibitem{Blei:07}
D.~Blei and J.~McAuliffe.
\newblock Supervised topic models.
\newblock In {\em NIPS}, 2007.

\bibitem{Blei:03}
D.~Blei, A.~Ng, and M.~I. Jordan.
\newblock {Latent Dirichlet Allocation}.
\newblock {\em JMLR}, (3):993--1022, 2003.

\bibitem{Bottou:1998}
L.~Bottou.
\newblock {\em Online Algorithms and Stochastic Approximations}.
\newblock Online Learning and Neural Networks, Edited by David Saad, Cambridge
  University Press, Cambridge, UK, 1998.

\bibitem{Bottou:2008}
L.~Bottou and O.~Bousquet.
\newblock The tradeoffs of large scale learning.
\newblock In {\em NIPS}, 2008.

\bibitem{Boyd:2011}
S.~Boyd, N.~Parikh, E.~Chu, B.~Peleato, and J.~Eckstein.
\newblock {\em Distributed Optimization and Statistical Learning via the
  Alternating Direction Method of Multipliers}, volume~3.
\newblock Foundations and Trends in Machine Learning, 2011.

\bibitem{Boyd:04}
S.~Boyd and L.~Vandenberghe.
\newblock {\em Convex Optimization}.
\newblock Cambridge University Press, 2004.

\bibitem{Brockwell:2006}
A.~Brockwell.
\newblock Parallel {Markov chain Monte Carlo} simulation by pre-fetching.
\newblock {\em JCGS}, 15(1):246--261, 2006.

\bibitem{Broderick:nips13}
T.~Broderick, N.~Boyd, A.~Wibisono, A.~C. Wilson, and M.~I. Jordan.
\newblock Streaming variational {Bayes}.
\newblock In {\em NIPS}, 2013.

\bibitem{Brumfiel:2011}
G.~Brumfiel.
\newblock High-energy physics: Down the petabyte highway.
\newblock {\em Nature}, 469:282--283, 2011.

\bibitem{Canny:2013}
J.~Canny and H.~Zhao.
\newblock Bidmach: Large-scale learning with zero memory allocation.
\newblock In {\em NIPS Big Learning Workshop}, 2013.

\bibitem{Chau:2013}
T.~Chau, J.~Targett, M.~Wijeyasinghe, W.~Luk, P.~Cheung, B.~Cope, A.~Eele, and
  J.~Maciejowski.
\newblock Accelerating sequential {Monte Carlo} method for real-time air
  traffic management.
\newblock {\em SIGARCH Computer Architecture News}, 41(5):35--40, 2013.

\bibitem{Chen:2016}
J.~Chen, K.~Li, J.~Zhu, and W.~Chen.
\newblock Warplda: a cache efficient o (1) algorithm for latent dirichlet
  allocation.
\newblock In {\em VLDB}, 2016.

\bibitem{Chen:2013}
J.~Chen, J.~Zhu, Z.~Wang, X.~Zheng, and B.~Zhang.
\newblock Scalable inference for logistic-normal topic models.
\newblock In {\em NIPS}, 2013.

\bibitem{Chen:icml14}
T.~Chen, E.~B. Fox, and C.~Guestrin.
\newblock Stochastic gradient {Hamiltonian Monte Carlo}.
\newblock In {\em ICML}, 2014.

\bibitem{Chu:2007}
C.~Chu, S.~K. Kim, Y.-A. Lin, Y.~Yu, G.~Bradski, A.~Y. Ng, and K.~Olukotun.
\newblock Map-reduce for machine learning on multicore.
\newblock In {\em NIPS}, 2007.

\bibitem{Crammer:2006}
K.~Crammer, O.~Dekel, J.~Keshet, S.~Shalev-Shwartz, and Y.~Singer.
\newblock Online passive-agressive algorithms.
\newblock {\em JMLR}, (7):551--585, 2006.

\bibitem{Dai:2013}
W.~Dai, J.~Wei, X.~Zheng, J.~K. Kim, S.~Lee, J.~Yin, Q.~Ho, and E.~Xing.
\newblock Petuum: A framework for iterative-convergent distributed {ML}.
\newblock In {\em arXiv:1312.7651}, 2013.

\bibitem{Dallaire:2014}
P.~Dallaire, P.~Giguere, and B.~Chaib-draa.
\newblock Learning the structure of probabilistic graphical models with an
  extended cascading {Indian} buffet process.
\newblock In {\em AAAI}, 2014.

\bibitem{Dean:2012}
J.~Dean, G.~Corrado, R.~Monga, K.~Chen, M.~Devin, M.~Mao, A.~Senior, P.~Tucker,
  K.~Yang, Q.~V. Le, et~al.
\newblock Large scale distributed deep networks.
\newblock In {\em NIPS}, 2012.

\bibitem{Dean:2004}
J.~Dean and S.~Ghemawat.
\newblock {MapReduce}: Simplified data processing on large clusters.
\newblock In {\em OSDI}, 2004.

\bibitem{Deng:nips11}
J.~Deng, S.~Satheesh, A.~Berg, and L.~Fei-Fei.
\newblock Fast and balanced: Efficient label tree learning for large scale
  object recognition.
\newblock In {\em NIPS}, 2011.

\bibitem{Doctorow:2008}
C.~Doctorow.
\newblock Big data: Welcome to the petacentre.
\newblock {\em Nature}, 455:16--21, 2008.

\bibitem{Donnet:2014}
S.~Donnet, V.~Rivoirard, J.~Rousseau, and C.~Scricciolo.
\newblock On convergence rates of empirical {Bayes} procedures.
\newblock In {\em 47th Scientific Meeting of the Italian Statistical Society},
  2014.

\bibitem{Doshi-Velez:2009}
F.~Doshi-Velez, K.~Miller, J.~V. Gael, and Y.~W. Teh.
\newblock Variational inference for the {Indian} buffet process.
\newblock In {\em AISTATS}, 2009.

\bibitem{Duan:2007}
J.~Duan, M.~Guindani, and A.~Gelfand.
\newblock Generalized spatial {Dirichlet} process models.
\newblock {\em Biometrika}, 94(4), 2007.

\bibitem{Duchi:2011}
J.~Duchi, E.~Hazan, and Y.~Singer.
\newblock Adaptive subgradient methods for online learning and stochastic
  optimization.
\newblock {\em JMLR}, 12:2121--2159, 2011.

\bibitem{DykMeng2001}
D.~V. Dyk and X.~Meng.
\newblock The art of data augmentation.
\newblock {\em JCGS}, 10(1):1--50, 2001.

\bibitem{Efron:science13}
B.~Efron.
\newblock Bayes' theorem in the 21st century.
\newblock {\em Science}, 340(6137):1177--1178, 2013.

\bibitem{Fan:NSR13}
J.~Fan, F.~Han, and H.~Liu.
\newblock Challenges of big data analysis.
\newblock {\em National Science Review}, 1(2):293--314, 2013.

\bibitem{Ferguson:73}
T.~Ferguson.
\newblock A {Bayesian} analysis of some nonparametric problems.
\newblock {\em Ann. Stats.}, (1):209--230, 1973.

\bibitem{Gal:parallDP}
Y.~Gal and Z.~Ghahramani.
\newblock Pitfalls in the use of parallel inference for the {Dirichlet}
  process.
\newblock In {\em ICML}, 2014.

\bibitem{Gelman:2013}
A.~Gelman, J.~Carlin, H.~Stern, D.~Dunson, A.~Vehtari, and D.~Rubin.
\newblock {\em Bayesian Data Analysis}.
\newblock Third Edition, Chapman \& Hall/CRC Texts in Statistical Science),
  2013.

\bibitem{Gelman:1992}
A.~Gelman and D.~Rubin.
\newblock Inference from iterative simulation using multiple simulations.
\newblock {\em Statist. Sci.}, 7(4):457--511, 1992.

\bibitem{Geman:1984}
S.~Geman and D.~Geman.
\newblock Stochastic relaxation, {Gibbs} distributions, and the {Bayesian}
  restoration of images.
\newblock {\em IEEE Trans. on PAMI}, 6(1):721--741, 1984.

\bibitem{George:2000}
E.~I. George and D.~P. Foster.
\newblock Calibration and empirical {Bayes} variable selection.
\newblock {\em Biometrika}, 87(4):731--747, 2000.

\bibitem{Gershman:2011}
S.~Gershman, P.~Frazier, and D.~Blei.
\newblock Distance dependent infinite latent feature models.
\newblock {\em arXiv:1110.5454}, 2011.

\bibitem{Gershmana:2012}
S.~Gershmana and D.~Blei.
\newblock A tutorial on {Bayesian} nonparametric models.
\newblock {\em J. Math. Psychol.}, (56):1--12, 2012.

\bibitem{Geyer:1995}
C.~J. Geyer and E.~A. Thompson.
\newblock Annealing {Markov} chain {Monte Carlo} with applications to ancestral
  inference.
\newblock {\em JASA}, 90(431):909--920, 1995.

\bibitem{Ghahramani:2013}
Z.~Ghahramani.
\newblock Bayesian nonparametrics and the probabilistic approach to modelling.
\newblock {\em Phil. Trans. of the Royal Society}, 2013.

\bibitem{Ghosh:book2003}
J.~K. Ghosh and R.~Ramamoorthi.
\newblock {\em Bayesian Nonparametrics}.
\newblock Springer, New York, NY, 2003.

\bibitem{Ghoting:2011}
A.~Ghoting, R.~Krishnamurthy, E.~Pednault, B.~Reinwald, V.~Sindhwani, and {et
  al}.
\newblock {SystemML}: Declarative machine learning on {MapReduce}.
\newblock In {\em ICDE}, 2011.

\bibitem{Gonzalez:2011}
J.~E. Gonzalez, Y.~Low, A.~Gretton, and C.~Guestrin.
\newblock Parallel {Gibbs} sampling: From colored fields to thin junction
  trees.
\newblock In {\em AISTATS}, 2011.

\bibitem{Gopalan:2013}
P.~Gopalan and D.~Blei.
\newblock Efficient discovery of overlapping communities in massive networks.
\newblock {\em PNAS}, 110(36):14534--14539, 2013.

\bibitem{Grelaud:2009}
A.~Grelaud, C.~P. Robert, J.-M. Marin, F.~Rodolphe, and J.-F. Taly.
\newblock Likelihood-free methods for model choice in {Gibbs} random fields.
\newblock {\em Bayesian Analysis}, 4(2):317--336, 2009.

\bibitem{Griffiths:nips06}
T.~Griffiths and Z.~Ghahramani.
\newblock Infinite latent feature models and the {Indian} buffet process.
\newblock In {\em NIPS}, 2006.

\bibitem{Griffiths:04}
T.~Griffiths and M.~Steyvers.
\newblock Finding scientific topics.
\newblock {\em PNAS}, 2004.

\bibitem{Guhaniyogi:2014}
R.~Guhaniyogi, S.~Qamar, and D.~Dunson.
\newblock Bayesian conditional density filtering for big data.
\newblock {\em arXiv:1401.3632}, 2014.

\bibitem{Hastings:1970}
W.~Hastings.
\newblock {Monte Carlo} sampling methods using {Markov} chains and their
  applications.
\newblock {\em Biometrika}, 57(1):97--109, 1970.

\bibitem{Hinton:2012}
G.~Hinton, L.~Deng, D.~Yu, A.~Mohamed, N.~Jaitly, and etc.
\newblock Deep neural networks for acoustic modeling in speech recognition.
\newblock {\em IEEE Signal Process. Mag.}, 29(6):82--97, 2012.

\bibitem{Hjort:2010}
N.~Hjort, C.~Holmes, P.~Muller, and S.~Walker.
\newblock {\em Bayesian Nonparametrics: Principles and Practice}.
\newblock Cambridge University Press, 2010.

\bibitem{Ho:2013}
Q.~Ho, J.~Cipar, H.~Cui, S.~Lee, J.~Kim, P.~Gibbons, G.~Gibson, G.~Ganger, and
  E.~Xing.
\newblock More effective distributed {ML} via a stale synchronous parallel
  parameter server.
\newblock In {\em NIPS}, 2013.

\bibitem{Hoffman:2013}
M.~D. Hoffman, D.~Blei, C.~Wang, and J.~Paisley.
\newblock Stochastic variational inference.
\newblock {\em JMLR}, 14:1303--1347, 2013.

\bibitem{Hofmann:2008}
T.~Hofmann, B.~Scholkopf, and A.~J. Smola.
\newblock Kernel methods in machine learning.
\newblock {\em Ann. Statist.}, 36(3):1171--1220, 2008.

\bibitem{Jasra:2007}
A.~Jasra, D.~A. Stephens, and C.~C. Holmes.
\newblock On population-based simulation for static inference.
\newblock {\em Statistics and Computing}, 17(3):263--279, 2007.

\bibitem{Jaynes:1968}
E.~T. Jaynes.
\newblock Prior probabilities.
\newblock {\em IEEE Trans. on Sys. Sci. and Cybernetics}, 4:227--241, 1968.

\bibitem{Jeffreys:1945}
H.~Jeffreys.
\newblock An invariant form for the prior probability in estimation problems.
\newblock {\em Proc. of the Royal Society of London. Series A, Mathematical and
  Physical Sciences}, 186(1007):453--461, 1945.

\bibitem{jia2014caffe}
Y.~Jia, E.~Shelhamer, J.~Donahue, S.~Karayev, J.~Long, R.~Girshick,
  S.~Guadarrama, and T.~Darrell.
\newblock Caffe: Convolutional architecture for fast feature embedding.
\newblock {\em arXiv preprint arXiv:1408.5093}, 2014.

\bibitem{Johnson:2013}
M.~J. Johnson, J.~Saunderson, and A.~S. Willsky.
\newblock Analyzing hogwild parallel gaussian {Gibbs} sampling.
\newblock In {\em NIPS}, 2013.

\bibitem{Jordan:2011}
M.~Jordan.
\newblock The era of big data.
\newblock {\em ISBA Bulletin}, 18(2):1--3, 2011.

\bibitem{Jordan:1999}
M.~Jordan, Z.~Ghahramani, T.~Jaakkola, and L.~Saul.
\newblock An introduction to variational methods for graphical models.
\newblock {\em MLJ}, 37(2):183--233, 1999.

\bibitem{Kadane:2004}
J.~B. Kadane and N.~A. Lazar.
\newblock Methods and criteria for model selection.
\newblock {\em JASA}, 99(465):279--290, 2004.

\bibitem{Kalman:1960}
R.~E. Kalman.
\newblock A new approach to linear filtering and prediction problems.
\newblock {\em J. Fluids Eng.}, 82(1):35--45, 1960.

\bibitem{Kass:1995}
R.~E. Kass and A.~E. Raftery.
\newblock Bayes factors.
\newblock {\em JASA}, 90(430):773--795, 1995.

\bibitem{Kim:2013}
D.~I. Kim, P.~Gopalan, D.~M. Blei, and E.~B. Sudderth.
\newblock Efficient online inference for {Bayesian} nonparametric relational
  models.
\newblock In {\em NIPS}, 2013.

\bibitem{Kingma:2014}
D.~Kingma and M.~Welling.
\newblock Auto-encoding variational bayes.
\newblock In {\em ICLR}, 2014.

\bibitem{Kingma:2014b}
D.~Kingma and M.~Welling.
\newblock Efficient gradient-based inference through transformations between
  {Bayes} nets and neural nets.
\newblock In {\em ICML}, 2014.

\bibitem{Korattikara:2014}
A.~Korattikara, Y.~Chen, and M.~Welling.
\newblock Austerity in {MCMC} land: Cutting the {Metropolis-Hastings} budget.
\newblock In {\em ICML}, 2014.

\bibitem{Koyejo:uai13}
O.~Koyejo and J.~Ghosh.
\newblock Constrained {Bayesian} inference for low rank multitask learning.
\newblock In {\em UAI}, 2013.

\bibitem{Kraska:2013}
T.~Kraska, A.~Talwalkar, J.~Duchi, R.~Griffith, M.~Franklin, and M.~Jordan.
\newblock {MLbase}: A distributed machine-learning system.
\newblock In {\em CIDR}, 2013.

\bibitem{Kyrola:2012}
A.~Kyrola, G.~E. Blelloch, and C.~Guestrin.
\newblock {GraphChi}: Large-scale graph computation on just a {PC}.
\newblock In {\em OSDI}, 2012.

\bibitem{Lake:BPL15}
B.~Lake, R.~Salakhutdinov, and J.~Tenenbaum.
\newblock Human-level concept learning through probabilistic program induction.
\newblock {\em Science}, 350(6266):1332--1338, 2015.

\bibitem{Lauritzen:1992}
S.~L. Lauritzen.
\newblock Propagation of probabilities, means and variances in mixed graphical
  association models.
\newblock {\em JASA}, 87:1098--1108, 1992.

\bibitem{Lawrence:2005}
N.~Lawrence.
\newblock Probabilistic non-linear principal component analysis with {Gaussian}
  process latent variable models.
\newblock {\em JMLR}, (6):1783--1816, 2005.

\bibitem{Quoc:2012}
Q.~Le, M.~Ranzato, R.~Monga, M.~Devin, K.~Chen, G.~Corrado, J.~Dean, and A.~Ng.
\newblock Building high-level features using large scale unsupervised learning.
\newblock In {\em ICML}, 2012.

\bibitem{LeCun-nature2015}
Y.~LeCun, Y.~Bengio, and G.~Hinton.
\newblock Deep learning.
\newblock {\em Nature}, 521:436--444, 2015.

\bibitem{Lee:2010}
A.~Lee, C.~Yau, M.~B. Giles, A.~Doucet, and C.~C. Holmes.
\newblock On the utility of graphics cards to perform massively parallel
  simulation of advanced {Monte Carlo} methods.
\newblock {\em JCGS}, 19(4):769--789, 2010.

\bibitem{Lee:2014}
S.~Lee, J.~K. Kim, X.~Zheng, Q.~Ho, G.~Gibson, and E.~Xing.
\newblock Primitives for dynamic big model parallelism.
\newblock {\em arXiv:1406.4580}, 2014.

\bibitem{Li:2014}
A.~Q. Li, A.~Ahmed, S.~Ravi, and A.~J. Smola.
\newblock Reducing the sampling complexity of topic models.
\newblock In {\em SIGKDD}, 2014.

\bibitem{Fei-Fei:05}
F.-F. Li and P.~Perona.
\newblock A bayesian hierarchical model for learning natural scene categories.
\newblock In {\em CVPR}, 2005.

\bibitem{Li:2016}
K.~Li, J.~Chen, W.~Chen, and J.~Zhu.
\newblock Saberlda: Sparsity-aware learning of topic models on gpus.
\newblock {\em arXiv preprint arXiv:1610.02496}, 2016.

\bibitem{Li:paramserv2014}
M.~Li, D.~Andersen, J.~W. Park, A.~Smola, A.~Ahmed, V.~Josifovski, J.~Long,
  E.~Shekita, and B.-Y. Su.
\newblock Scaling distributed machine learning with the parameter server.
\newblock In {\em OSDI}, 2014.

\bibitem{Liu:1998}
J.~S. Liu and R.~Chen.
\newblock Sequential {Monte Carlo} methods for dynamic systems.
\newblock {\em JASA}, 93(443):1032--1044, 1998.

\bibitem{Liu:2011}
Z.~Liu, Y.~Zhang, E.~Y. Chang, and M.~Sun.
\newblock {PLDA+}: Parallel latent {Dirichlet} allocation with data placement
  and pipeline processing.
\newblock {\em TIST}, 2(3), 2011.

\bibitem{Lloyd:2014}
J.~Lloyd, D.~Duvenaud, R.~Grosse, J.~Tenenbaum, and Z.~Ghahramani.
\newblock Automatic construction and natural language description of
  nonparametric regression models.
\newblock In {\em AAAI}, 2014.

\bibitem{Low:2010}
Y.~Low, J.~Gonzalez, A.~Kyrola, D.~Bickson, C.~Guestrin, and J.~Hellerstein.
\newblock Graphlab: A new framework for parallel machine learning.
\newblock In {\em UAI}, 2013.

\bibitem{MacEachern:1999}
S.~MacEachern.
\newblock Dependent nonparametric processes.
\newblock In {\em ASA proceedings of the section on Bayesian statistical
  science}, 1999.

\bibitem{Maclaurin:2014}
D.~Maclaurin and R.~P. Adams.
\newblock Firefly {Monte Carlo}: Exact {MCMC} with subsets of data.
\newblock In {\em UAI}, 2014.

\bibitem{Malewicz:2010}
G.~Malewicz, M.~H. Austern, A.~J. Bik, J.~C. Dehnert, I.~Horn, N.~Leiser, and
  G.~Czajkowski.
\newblock Pregel: a system for large-scale graph processing.
\newblock In {\em SIGMOD}, 2010.

\bibitem{Mandt:2014}
S.~Mandt and D.~Blei.
\newblock Smoothed gradients for stochastic variational inference.
\newblock {\em arXiv:1406.3650}, 2014.

\bibitem{Marlin:2011}
B.~Marlin, E.~Khan, and K.~Murphy.
\newblock Piecewise bounds for estimating {Bernoulli}-logistic latent
  {Gaussian} models.
\newblock In {\em ICML}, 2011.

\bibitem{McAuliffe:2006}
J.~D. McAuliffe, D.~M. Blei, and M.~I. Jordan.
\newblock Nonparametric empirical {Bayes} for the {Dirichlet} process mixture
  model.
\newblock {\em Statistics and Computing}, 16(1):5--14, 2006.

\bibitem{Mei:robustBayes14}
S.~Mei, J.~Zhu, and X.~Zhu.
\newblock Robust {RegBayes}: Selectively incorporating first-order logic domain
  knowledge into {Bayesian} models.
\newblock In {\em ICML}, 2014.

\bibitem{Metropolis:1953}
N.~Metropolis, A.~Rosenbluth, M.~Rosenbluth, A.~Teller, and E.~Teller.
\newblock Equation of state calculations by fast computing machines.
\newblock {\em J. Chem. Phys}, 21(6):1087, 1953.

\bibitem{Miller:nips09}
K.~Miller, T.~Griffiths, and M.~Jordan.
\newblock Nonparametric latent feature models for link prediction.
\newblock In {\em NIPS}, 2009.

\bibitem{Minsker:2014}
S.~Minsker, S.~Srivastava, L.~Lin, and D.~B. Dunson.
\newblock Scalable and robust {Bayesian} inference via the median posterior.
\newblock In {\em ICML}, 2014.

\bibitem{Mitchell:1997}
T.~Mitchell.
\newblock {\em Machine Learning}.
\newblock McGraw Hill, 1997.

\bibitem{Mnih:2014}
A.~Mnih and K.~Gregor.
\newblock Neural variational inference and learning in belief networks.
\newblock In {\em ICML}, 2014.

\bibitem{Moral:2006}
P.~D. Moral, A.~Doucet, and A.~Jasra.
\newblock Sequential {Monte Carlo} samplers.
\newblock {\em J. R. Stat. Soc., Ser B}, 68(3):411--436, 2006.

\bibitem{Muller:2004}
P.~Muller and F.~A. Quintana.
\newblock Nonparametric {Bayesian} data analysis.
\newblock {\em Statistical Science}, 19(1):95--110, 2004.

\bibitem{Neal:2000}
R.~Neal.
\newblock Markov chain sampling methods for {Dirichlet} process mixture models.
\newblock {\em JCGS}, pages 249--265, 2000.

\bibitem{Neal:slice2003}
R.~Neal.
\newblock Slice sampling.
\newblock {\em Ann. Statist.}, 31(3):705--767, 2003.

\bibitem{Neal:10}
R.~Neal.
\newblock {\em {MCMC} using {Hamiltonian} Dynamics}.
\newblock Handbook of Markov Chain Monte Carlo (S. Brooks, A. Gelman, G. Jones,
  and X.-L. Meng, eds.), Chapman \& Hall / CRC Press, 2010.

\bibitem{Neiswanger:2014}
W.~Neiswanger, C.~Wang, and E.~P. Xing.
\newblock Asymptotically exact, embarrassingly parallel {MCMC}.
\newblock In {\em UAI}, 2014.

\bibitem{Newman:2007}
D.~Newman, A.~Asuncion, P.~Smyth, and M.~Welling.
\newblock Distributed inference for latent {Dirichlet} allocation.
\newblock In {\em NIPS}, 2007.

\bibitem{Niu:2011}
F.~Niu, B.~Recht, C.~Re, and S.~J. Wright.
\newblock Hogwild: A lock-free approach to parallelizing stochastic gradient
  descent.
\newblock In {\em NIPS}, 2011.

\bibitem{Opper:1999}
M.~Opper.
\newblock {\em A {Bayesian} Approach to Online Learning}.
\newblock {\it Online Learning in Neural Networks}, Cambridge University, 1999.

\bibitem{Paisley:2012}
J.~Paisley, D.~M. Blei, and M.~I. Jordan.
\newblock Variational {Bayesian} inference with stochastic search.
\newblock In {\em ICML}, 2012.

\bibitem{Papaspiliopoulos:2007}
O.~Papaspiliopoulos, G.~O. Roberts, and M.~Skold.
\newblock A general framework for the parametrization of hierarchical models.
\newblock {\em Statistical Science}, 22(1):59--73, 2007.

\bibitem{Teh:nips13}
S.~Patterson and Y.~W. Teh.
\newblock Stochastic gradient {Riemannian Langevin} dynamics on the probability
  simplex.
\newblock In {\em NIPS}, 2013.

\bibitem{Petrone:2014}
S.~Petrone, J.~Rousseau, and C.~Scricciolo.
\newblock Bayes and empirical {Bayes}: do they merge?
\newblock {\em Biometrika}, pages 1--18, 2014.

\bibitem{Pillai:2014}
N.~Pillai and A.~Smith.
\newblock Ergodicity of approximate {MCMC} chains with applications to large
  data sets.
\newblock {\em arXiv:1405.0182}, 2014.

\bibitem{Pitman:2002}
J.~Pitman.
\newblock Combinatorial stochastic processes.
\newblock {\em Technical Report No. 621. Department of Statistics, UC,
  Berkeley}, 2002.

\bibitem{Power:2010}
R.~Power and J.~Li.
\newblock Piccolo: Building fast, distributed programs with partitioned tables.
\newblock In {\em OSDI}, 2010.

\bibitem{Rabiner:89}
L.~R. Rabiner.
\newblock A tutorial on hidden {Markov} models and selected applications in
  speech recognition.
\newblock {\em Proc. of the IEEE}, 77(2):257--286, 1989.

\bibitem{Ranganath:2013}
R.~Ranganath, C.~Wang, D.~Blei, and E.~Xing.
\newblock An adaptive learning rate for stochastic variational inference.
\newblock In {\em ICML}, 2013.

\bibitem{Rasmussen:2006}
C.~E. Rasmussen and C.~K.~I. Williams.
\newblock {\em Gaussian Processes for Machine Learning}.
\newblock The MIT Press, 2006.

\bibitem{Reichman:2011}
O.~Reichman, M.~Jones, and M.~Schildhauer.
\newblock Challenges and opportunities of open data in ecology.
\newblock {\em Science}, 331(6018):703--705, 2011.

\bibitem{Rezende:2014}
D.~J. Rezende, S.~Mohamed, and D.~Wierstra.
\newblock Stochastic backpropagation and approximate inference in deep
  generative models.
\newblock In {\em ICML}, 2014.

\bibitem{Robert:2005}
C.~Robert and G.~Casella.
\newblock {\em {Monte Carlo} Statistical Methods}.
\newblock Springer, 2005.

\bibitem{Robert:2011}
C.~P. Robert, J.-M. Cornuet, J.-M. Marin, and N.~S. Pillai.
\newblock Lack of confidence in approximate {Bayesian} computation model
  choice.
\newblock {\em PNAS}, 108(37):15112--15117, 2011.

\bibitem{Roberts:02}
G.~Roberts and O.~Strame.
\newblock Langevin diffusions and {Metropolis-Hastings} algorithms.
\newblock {\em Methodology and Computing in Applied Probability}, 4:337--357,
  2002.

\bibitem{Huelsenbeck:2003}
F.~Ronquist and J.~Huelsenbeck.
\newblock {MrBayes: Bayesian} inference of phylogenetic trees.
\newblock {\em Bioinformatics}, 19(12):1572--1574, 2003.

\bibitem{Salakhutdinov:2009}
R.~Salakhutdinov.
\newblock Learning deep generative models.
\newblock {\em PhD Thesis, University of Toronto}, 2009.

\bibitem{Salihoglu:gps2013}
S.~Salihoglu and J.~Widom.
\newblock {GPS}: A graph processing system.
\newblock In {\em SSDBM}, 2013.

\bibitem{Schraudolph:2007}
N.~Schraudolph, J.~Yu, and S.~Gunter.
\newblock A stochastic {Quasi-Newton} method for online convex optimization.
\newblock In {\em AISTATS}, 2007.

\bibitem{Scott:2013}
S.~Scott, A.~Blocker, F.~Bonassi, H.~Chipman, E.~George, and R.~McCulloch.
\newblock Bayes and big data: The consensus {Monte Carlo} algorithm.
\newblock {\em EFaB Bayes 250 Workshop}, 16, 2013.

\bibitem{Scott:2002}
S.~L. Scott.
\newblock Bayesian methods for hidden {Markov} models.
\newblock {\em JASA}, 97(457):337--351, 2002.

\bibitem{Sethuraman:94}
J.~Sethuraman.
\newblock A constructive definition of dirichlet priors.
\newblock {\em Statistica Sinica}, (4):639--650, 1994.

\bibitem{Shi:2014}
T.~Shi and J.~Zhu.
\newblock Online {Bayesian} passive-aggressive learning.
\newblock In {\em ICML}, 2014.

\bibitem{Smola:vldb10}
A.~Smola and S.~Narayanamurthy.
\newblock An architecture for parallel topic models.
\newblock {\em VLDB}, 2010.

\bibitem{Snoek:2012}
J.~Snoek, H.~Larochelle, and R.~P. Adams.
\newblock Practical {Bayesian} optimization of machine learning algorithms.
\newblock In {\em NIPS}, 2012.

\bibitem{Srivastava:2014}
N.~Srivastava, G.~Hinton, A.~Krizhevsky, I.~Sutskever, and R.~Salakhutdinov.
\newblock Dropout: A simple way to prevent neural networks from overfitting.
\newblock {\em JMLR}, 15:1929--1958, 2014.

\bibitem{Strid:2010}
I.~Strid.
\newblock Efficient parallelisation of {Metropolis-Hastings} algorithms using a
  prefetching approach.
\newblock {\em Computational Statistics and Data Analysis}, 54:2814--2835,
  2010.

\bibitem{Suchard:2010}
M.~Suchard, Q.~Wang, C.~Chan, J.~Frelinger, A.~Cron, and M.~West.
\newblock Understanding {GPU} programming for statistical computation: Studies
  in massively parallel massive mixtures.
\newblock {\em JCGS}, 19(2):419--438, 2010.

\bibitem{Tan:jmlr14}
M.~Tan, I.~Tsang, and L.~Wang.
\newblock Towards ultrahigh dimensional feature selection for big data.
\newblock {\em JMLR}, (15):1371--1429, 2014.

\bibitem{Tanner:1987}
M.~Tanner and W.~Wong.
\newblock The calculation of posterior distributions by data augmentation.
\newblock {\em JASA}, 82(398):528--540, 1987.

\bibitem{Teh:2007}
Y.~W. Teh, D.~Gorur, and Z.~Ghahramani.
\newblock Stick-breaking construction for the {Indian} buffet process.
\newblock In {\em AISTATS}, 2007.

\bibitem{Teh:2006}
Y.~W. Teh, M.~I. Jordan, M.~J. Beal, and D.~M. Blei.
\newblock Hierarchical {Dirichlet} processes.
\newblock {\em JASA}, 101(476):1566--1581, 2006.

\bibitem{Teh:2014sgld}
Y.~W. Teh, A.~Thiey, and S.~Vollmer.
\newblock Consistency and fluctuations for stochastic gradient {Langevin}
  dynamics.
\newblock {\em arXiv:1409.0578}, 2014.

\bibitem{2016arXiv160502688short}
{Theano Development Team}.
\newblock {Theano: A {Python} framework for fast computation of mathematical
  expressions}.
\newblock {\em arXiv e-prints}, abs/1605.02688, May 2016.

\bibitem{Thibaux:2007}
R.~Thibaux and M.~I. Jordan.
\newblock Hierarchical {Beta} processes and the {Indian} buffet process.
\newblock In {\em AISTATS}, 2007.

\bibitem{Turnera:2012}
B.~M. Turnera and T.~V. Zandtb.
\newblock A tutorial on approximate {Bayesian} computation.
\newblock {\em J. Math. Psychol.}, 56(2):69--85, 2012.

\bibitem{pcgs}
D.~van Dyk and T.~Park.
\newblock Partially collapsed {Gibbs} samplers: Theory and methods.
\newblock {\em JASA}, 103(482):790--796, 2008.

\bibitem{Vincent:2008}
P.~Vincent, H.~Larochelle, Y.~Bengio, and P.-A. Manzagol.
\newblock Extracting and composing robust features with denoising autoencoders.
\newblock In {\em ICML}, 2008.

\bibitem{Wainwright:2008}
M.~Wainwright and M.~Jordan.
\newblock Graphical models, exponential families, and variational inference.
\newblock {\em Foundations and Trends in Machine Learning}, 1(1--2):1--305,
  2008.

\bibitem{Walker:2007}
S.~G. Walker.
\newblock Sampling the {Dirichlet} mixture model with slices.
\newblock {\em Commun Stat - Simul and Comput}, 36:45--54, 2007.

\bibitem{Wang:2013}
X.~Wang and D.~B. Dunson.
\newblock Parallelizing {MCMC} via {Weierstrass} sampler.
\newblock {\em arXiv:1312.4605}, 2013.

\bibitem{Wang:2014towards}
Y.~Wang, X.~Zhao, Z.~Sun, H.~Yan, L.~Wang, Z.~Jin, L.~Wang, Y.~Gao, J.~Zeng,
  Q.~Yang, et~al.
\newblock Towards topic modeling for big data.
\newblock {\em arXiv:1405.4402}, 2014.

\bibitem{Weinberger:2009}
K.~Weinberger, A.~Dasgupta, J.~Langford, A.~Smola, and J.~Attenberg.
\newblock Feature hashing for large scale multitask learning.
\newblock In {\em ICML}, 2009.

\bibitem{Welling:2014}
M.~Welling.
\newblock Exploiting the statistics of learning and inference.
\newblock In {\em NIPS workshop on "Probabilistic Models for Big Data"}, 2013.

\bibitem{Welling:icml11}
M.~Welling and Y.~W. Teh.
\newblock Bayesian learning via stochastic gradient {Langevin} dynamics.
\newblock In {\em ICML}, 2011.

\bibitem{Wilkinson:2004}
D.~Wilkinson.
\newblock Parallel {Bayesian} computation.
\newblock {\em Handbook of Parallel Computing and Statistics, Chapter 18},
  2004.

\bibitem{Williams:BayesCond1980}
P.~M. Williams.
\newblock Bayesian conditionalisation and the principle of minimum information.
\newblock {\em The British Journal for the Philosophy of Science}, 31(2), 1980.

\bibitem{Williamson:2010}
S.~Williamson, P.~Orbanz, and Z.~Ghahramani.
\newblock Dependent {Indian} buffet processes.
\newblock In {\em AISTATS}, 2010.

\bibitem{Williamson:2013}
S.~A. Williamson, A.~Dubey, and E.~P. Xing.
\newblock Parallel {Markov} chain {Monte Carlo} for nonparametric mixture
  models.
\newblock In {\em ICML}, 2013.

\bibitem{Wu:2012}
X.-L. Wu, C.~Sun, T.~Beissinger, G.~Rosa, K.~Weigel, N.~Gatti, and D.~Gianola.
\newblock Parallel {Markov chain Monte Carlo} - bridging the gap to
  high-performance {Bayesian} computation in animal breeding and genetics.
\newblock {\em Genetics Seletion Evolution}, 44(1):29--46, 2012.

\bibitem{Xin:2013}
R.~S. Xin, J.~E. Gonzalez, M.~J. Franklin, and I.~Stoica.
\newblock Graphx: A resilient distributed graph system on spark.
\newblock In {\em Workshop on Graph Data Management Experiences and Systems},
  2013.

\bibitem{Minjie:2014}
M.~Xu, B.~Lakshminarayanan, Y.~Teh, J.~Zhu, and B.~Zhang.
\newblock Distributed {Bayesian} posterior sampling via moment sharing.
\newblock In {\em NIPS}, 2014.

\bibitem{Yan:2009}
F.~Yan, N.~Xu, and A.~Qi.
\newblock Parallel inference for latent {Dirichlet} allocation on graphics
  processing units.
\newblock In {\em NIPS}, 2009.

\bibitem{Yang:KDD16}
Y.~Yang, J.~Chen, and J.~Zhu.
\newblock Distributing the stochastic gradient sampler for large-scale lda.
\newblock In {\em KDD}, 2016.

\bibitem{YuMeng:2011}
Y.~Yu and X.-L. Meng.
\newblock To center or not to center: That is not the question--an
  ancillarity--sufficiency interweaving strategy (asis) for boosting mcmc
  efficiency.
\newblock {\em JCGS}, 20(3):531--570, 2011.

\bibitem{Zaharia:2010}
M.~Zaharia, M.~Chowdhury, M.~J. Franklin, S.~Shenker, and I.~Stoica.
\newblock Spark: cluster computing with working sets.
\newblock In {\em Hot Topics in Cloud Computing}, 2010.

\bibitem{Zhai:2012}
K.~Zhai, J.~Boyd-Graber, N.~Asadi, and M.~L. Alkhouja.
\newblock {Mr. LDA}: a flexible large scale topic modeling package using
  variational inference in {MapReduce}.
\newblock In {\em WWW}, 2012.

\bibitem{Zhang:2014}
A.~Zhang, J.~Zhu, and B.~Zhang.
\newblock Max-margin infinite hidden {Markov} models.
\newblock In {\em ICML}, 2014.

\bibitem{Zheng:arxiv2014}
X.~Zheng, J.~K. Kim, Q.~Ho, and E.~Xing.
\newblock Model-parallel inference for big topic models.
\newblock {\em arXiv:1411.2305}, 2014.

\bibitem{Zhu:ICML12}
J.~Zhu.
\newblock Max-margin nonparametric latent feature models for link prediction.
\newblock In {\em ICML}, 2012.

\bibitem{Zhu:jmlr12}
J.~Zhu, A.~Ahmed, and E.~Xing.
\newblock {MedLDA}: maximum margin supervised topic models.
\newblock {\em JMLR}, (13):2237--2278, 2012.

\bibitem{Zhu:ICML13}
J.~Zhu, N.~Chen, H.~Perkins, and B.~Zhang.
\newblock Gibbs max-margin topic models with fast sampling algorithms.
\newblock In {\em ICML}, 2013.

\bibitem{Zhu:regBayes14}
J.~Zhu, N.~Chen, and E.~Xing.
\newblock Bayesian inference with posterior regularization and applications to
  infinite latent {SVMs}.
\newblock {\em JMLR}, 15:1799--1847, 2014.

\bibitem{Zhu:ACL13}
J.~Zhu, X.~Zheng, and B.~Zhang.
\newblock Improved {Bayesian} logistic supervised topic models with data
  augmentation.
\newblock In {\em ACL}, 2013.

\bibitem{Zhu:KDD13}
J.~Zhu, X.~Zheng, L.~Zhou, and B.~Zhang.
\newblock Scalable inference in max-margin supervised topic models.
\newblock In {\em SIGKDD}, 2013.

\bibitem{Zinkevich:2010}
M.~A. Zinkevich, M.~Weimer, A.~Smola, and L.~Li.
\newblock Parallelized stochastic gradient descent.
\newblock In {\em NIPS}, 2010.

\end{thebibliography}

\end{document}